%% file: mol2015.tex
\documentclass[11pt]{article}
\usepackage{acl2012}
\usepackage{times}
\usepackage{latexsym}
\usepackage{amsmath}
\usepackage{amssymb}
\usepackage{amsfonts}

\usepackage[multiple]{footmisc}

\usepackage{multirow}
\usepackage{url}

\usepackage{tikz}
\usepackage{ccg}
\usepackage{gb4e}
\usepackage{lipsum}

\usepackage{stmaryrd}

\usepackage{tipa}

\title{A Frobenius Model of Information Structure\\in Categorical Compositional Distributional Semantics}

\author{Dimitri Kartsaklis\quad\quad\quad\quad\quad Mehrnoosh Sadrzadeh\vspace{0.2cm} \\ 
    Queen Mary University of London\\School of Electronic Engineering and Computer Science\\
    Mile End Road, London E1 4NS, UK\\
    {\small {\tt \{d.kartsaklis;m.sadrzadeh\}@qmul.ac.uk}}}

\input{macros}
\input{colours}

\begin{document}

\maketitle


\begin{abstract}
The categorical compositional distributional model of \newcite{Coeckeetal} provides a linguistically motivated procedure for computing the meaning of a sentence as a function of the distributional meaning of the words therein. The theoretical framework allows for reasoning about compositional aspects of language and offers structural ways of studying the underlying relationships. While the model so far has been applied on the level of syntactic structures, a sentence can bring extra information  conveyed in utterances via intonational means. In the current paper we extend the framework in order to accommodate this additional information, using Frobenius algebraic structures canonically induced over the basis of finite-dimensional vector spaces.  We detail the theory, provide truth-theoretic and distributional semantics for meanings of intonationally-marked utterances, and present justifications and extensive examples.
\end{abstract}

\section{Introduction}

Distributional models of meaning, in which a word is represented as a high dimensional vector of contextual statistics in a metric space, provide a convincing framework for lexical semantics that has been found useful in a number of natural language processing tasks \cite{Schutze,Landauer,Manning}. Despite their success at the word level, the underlying hypothesis  of these approaches does not naturally scale up to phrases or sentences due to the infinite capacity of language to produce new meanings from a finite vocabulary and a set of grammar rules.

\newcite{Coeckeetal} provide a solution to the problem by noticing that the category of finite-dimensional vector spaces and linear maps is homomorphic to a grammar expressed as a pregroup \cite{Lambek}; specifically, both share compact closed structure \cite{kelly1972many}. In practice this means that any grammatical derivation based on the type-logical identities of the individual words in a sentence can be translated to a (multi-)linear map which, when applied on the vectorial representations of the words therein, results in a sentence vector. The grammatical type of a word determines the vector space in which this word lives. Taking nouns to be simple vectors in a basic vector space $N$, an adjective, for example, becomes a linear map $N\to N$, or equivalently, a matrix in $N\ten N$; furthermore, a transitive verb is a bi-linear map $N\ten N \to S$, living in $N\ten S \ten N$. Composition takes the form of tensor contraction, which is a generalization of matrix multiplication to higher order tensors. 

In general, the model resembles a quantitative linear-algebraic version of the formal semantics approach \cite{Mon1}, in the sense that syntax strictly guides the semantic composition. Interestingly, syntax seems to co-exist with a distinct structural layer, the purpose of which is to optimize the message that an utterance conveys. This aspect is known as \textit{information structure}, and at the phrase or sentence level is expressed as a distinction between a theme part (information that is generally agreed to be known to both of the interlocutors) and a rheme part---information that is new for the addressee. The exact relation that holds between syntactical and information structure is an interesting and controversial topic. For example, a theme does not have to comprise a valid grammatical constituent in the strict sense of the term, as it is evident in the following example:

\begin{exe}
  \ex Q: Do you need anything? \\
      A: \theme{I would like} \rheme{some tea}
\end{exe}

The distinction between a theme and rheme is denoted by the presence of a boundary that can be expressed by phonological, morphological or even syntactical means, depending on the language. Furthermore, the presence of such boundaries suggest the existence of a distinct composition operator related to information structure and different than the one that would be normally used for syntax. 

In this paper we extend the categorical model of \newcite{Coeckeetal} in a way to accommodate an information structure layer of composition. In order to achieve this, we model intonational boundaries (the devices for defining information structure in English) by using the multiplication part of the Frobenius algebra that is canonically induced over any vector space with fixed basis, in order to endow equal contribution of the theme and rheme on the vectorial representation of a sentence, thus putting emphasis on the appropriate part. The resulting model can be seen as containing two types of composition operators: the usual tensor contraction for accommodating syntax, and the Frobenius multiplication for accommodating information structure. We discuss the implications in terms of the resulting vectorial representations for phrases and sentences, and provide connections with existing models from the current literature of compositional distributional semantics. Various examples demonstrate the potential of the model.

\section{Categorical compositional distributional semantics}

The categorical model of \newcite{Coeckeetal} assigns semantic representations to phrases and sentences of language, based on their grammatical structure and the semantics of individual words. In its most abstract form, this model can be expressed in terms of a structure-preserving passage between grammar and meaning:

\begin{equation*}
\F \colon \textmd{Grammar} \to \textmd{Meaning}
\end{equation*}

Given a sequence of words $\textmd{w}_1\cdots\textmd{w}_n$, its categorical meaning is defined to be:


\small
\begin{equation}
\label{gendef}
\sem{{\textmd{w}_1}\cdots{\textmd{w}_n}} := \F(\alpha)(\sem{\textmd{w}_1},\cdots,\sem{\textmd{w}_n})
\end{equation}
\normalsize

Here, $\alpha$ is derived from the  grammatical relationships amongst the words in the sequence. This notion can be formalised in a coherent way, if both the grammar and the meaning are expressed in a high level logical structure, referred to by \emph{compact closure}. Lambek's pregroup algebras \cite{Lambek} and vector space distributional semantics are examples of compact closed structures. Stipulating that the grammar is expressed in a pregroup algebra and that the meaning of words are vectors constructed using the distributional hypothesis \cite{harris}, Eq. \ref{gendef} gets a more concrete form:\footnote{One can translate the types from other type logics, such as the syntactic calculus and CCG to pregroups and carry on with the same calculations. There is also recent work that directly assigns vector semantics to CCG \cite{LewisSteedman}.}

\small
\begin{equation}
\label{specdef}
\ov{\textmd{w}_1\cdots\textmd{w}_n}:= 
\F(\alpha)(\ov{\textmd{w}}_1 \otimes \cdots \otimes \ov{\textmd{w}}_n)
\end{equation}
\normalsize

In the proceeding subsections we make these notions precise and provide intuitions and examples. 

\subsection{Pregroup grammars}
\label{sec:preg-grammar}
A pregroup grammar is a pregroup algebra,  linked to the vocabulary of a language  via the notion of  a type dictionary.  We define these structures below. 

A pregroup algebra is a partially ordered monoid where each element has a left and a right adjoint. It is denoted by a tuple $(P, \leq, \cdot, 1, (-)^l, (-)^r)$, where $(P, \leq)$ is a partially ordered set, and $\cdot$ is a monoid multiplication with 1 as its unit. For each element $p \in P$ there are $p^l, p^r \in P$, referred to by $p$'s \emph{left} and \emph{right} adjoints, satisfying the following four inequalities:

\vspace{-0.5cm}
\begin{equation*}
p \cdot p^r \leq 1 \leq p^r \cdot p 
\qquad
p^l \cdot p \leq 1 \leq p \cdot p^l
\end{equation*}
\vspace{-0.5cm}

When a pregroup algebra is generated over a base set ${\cal B}$, it is denoted by $P({\cal B})$.  Given the vocabulary of a language $\Sigma$  and a set of its basic grammatical types ${\cal B}$, a pregroup grammar is a relation $D \subseteq \Sigma  \times P({\cal B})$ that assigns grammatical types from the pregroup algebra  $P({\cal B})$ to the words of the vocabulary $\Sigma$.  Such a pregroup grammar is denoted by  $P({\cal B}, \Sigma)$. 

As an example, suppose ${\cal B} = \{n, s\}$, where $n$ stands for a well-formed noun phrase and $s$ for a well-formed sentence. Suppose further that $\Sigma = \{\textmd{Mary}, \textmd{snores}, \textmd{likes}, \textmd{musicals}\}$. The pregroup dictionary consists of the following  set:

\small
\begin{equation*}
\left \{(\textmd{Mary}, n),  (\textmd{snores}, n^r  s), (\textmd{likes}, n^r  s  n^l), (\textmd{musicals}, n)\right\}
\end{equation*}
\normalsize

One says that a  sequence of words $\textmd{w}_1\textmd{w}_2\cdots \textmd{w}_n$ for $\textmd{w}_i \in \Sigma$  forms a grammatical sentence, according to a pregroup grammar $P({\cal B}, \Sigma)$, whenever we have:

\begin{equation*}
t_1 \cdot t_2 \cdot \hdots \cdot t_n \leq s
\end{equation*}

\noindent
for $(\textmd{w}_i, t_i) \in D$. The above inequality is often referred to by  \emph{grammatical reduction}. For example, `Mary likes musicals' is a grammatical sentence, since we have the following reduction:

\begin{equation}
n \cdot n^r \cdot s \cdot  n^l \cdot n \leq 1 \cdot s \cdot 1 = s
\label{equ:reduction}
\end{equation}

\subsection{Distributional models}

The only piece of information provided by a derivation like the one in Eq. \ref{equ:reduction} is whether the sentence in question is well-formed or not. Furthermore, we are unable to distinguish between words of the same type. Distributional models of meaning provide a solution to these problems by following the \textit{distributional hypothesis }\cite{harris}, which states that semantically similar words must appear in similar contexts. Hence, the semantic representation of a word can be given in terms of its distributional behaviour in a large corpus of text. In its simplest form, a word vector is comprised by numbers that show how many times the target word co-occurs with every other word in a selected subset of the vocabulary (usually the most frequent content-bearing words). This allows the representation of words as points in some high dimensional space, where semantic relatedness can be measured (usually by cosine distance) and evaluated. For a concise introduction to distributional models, see \cite{turney2010}. We will now proceed to show how the quantitative approach of distributional models can be combined with the compositional model of Section \ref{sec:preg-grammar} into a unified account.

\subsection{Categorical generalization}

The theory of categories generalises algebraic constructions to categorical ones \cite{maclane}. Herein,  instead of sets, functions or relations, one has objects $A, B$ and morphisms $f \colon A \to B$.  The generalised binary operation over these is referred to by a product. Posing different conditions on the objects, morphisms, or the product results in different kinds of categories. A \textit{monoidal category} has a product with a unit $I$, that is $A \otimes I \cong I \otimes A \cong A$. These categories are generalisations of partially ordered monoids: elements of the partial order become objects of the category and the partial orderings between them become  morphisms.  
Furthermore, \textit{compact closed categories} are generalisations of pregroups, where the adjunction inequalities correspond to the following $\epsilon$ and $\eta$ morphisms:

\vspace{-0.3cm}
\begin{eqnarray*}
\epsilon^r \colon A \otimes A^r \to I &\qquad& \eta^r \colon I \to A^r \otimes A \\
\epsilon^l \colon A^l \otimes A \to I &\qquad& \eta^l \colon I \to A \otimes A^l
\end{eqnarray*}
\vspace{-0.3cm}

These maps needs to adhere to four axioms, referred to as \textit{yanking equations}, which ensure that all relevant diagrams commute.


The importance of the theory of categories for this paper is that finite-dimensional vector spaces and linear maps also form a compact closed category, denoted by $\mathbf{FVect}$. Herein, objects are vector spaces, morphisms are linear maps, and the product is the tensor product between vector spaces whose unit is the scalar field of the vector spaces, in our case, real numbers ($\mathbb{R}$). In the presence of a fixed basis (which is the case we are interested in) the adjoints become identity, that is we have $V^r \cong V^l \cong V$, for a vector space $V$ spanned by $\{\ov{v}_i\}_i$. As a result the four $\epsilon$ and $\eta$ maps reduce to two:

\begin{equation*}
\epsilon \colon V \otimes V \to \mathbb{R} \qquad
\eta \colon \mathbb{R} \to V \otimes V
\end{equation*}

The $\epsilon$ map takes the inner product of two vectors and the $\eta$ map produces a diagonal matrix. The fact that both pregroup algebras and vector spaces form compact closed categories allows us to develop a structure preserving passage between the two mathematical structures, thus enabling us to bridge the grammatical structure to distributional semantics. 

\subsection{From grammar to distributions}

A structure preserving passage from grammatical structures (in the form of a pregroup grammar) to semantics  (in the form of vector spaces)  is given by a map  denoted as follows:

\begin{equation}
  \mathcal{F}:P(\Sigma, {\cal B}) \to \mathbf{FVect}
  \label{equ:functor}
\end{equation}

This is a strongly monoidal passage, which means that it has the following compositional properties for juxtapositions of types in a pregroup grammar:

\small
\begin{eqnarray}
\label{equ:m1}\mathcal{F}  (1) &=& \mathbb{R}\\
\label{equ:m2}\mathcal{F} (p \cdot q) &=&   \mathcal{F}(p) \otimes   \mathcal{F}(q)\\
\label{equ:m3}\mathcal{F} (p^r) = \mathcal{F} (p^l) &=&   \mathcal{F} (p)\\
\label{equ:m4}\mathcal{F} (p \leq q) &=&   \mathcal{F} (p) \to \mathcal{F} (q)    
\end{eqnarray}
\normalsize

%
On the level of basic types we assign a vector space to each basic type, that is,   $\mathcal{F}(n) = N$ and  $\mathcal{F} (s) = S$.  
As a result of the above assignments,  words that have simple types, for example noun phrases, will become vectors in vector space $N$. Words that are functions of one argument become matrices, e.g. intransitive verbs with type $n^r\cdot s$ are elements of $N \otimes S$; and words that are functions of two arguments, e.g. transitive verbs with type $n^r \cdot s \cdot  n^l$, become  tensors of order 3, living in $N \otimes S \otimes N$ for the specific case.  The grammatical reductions are translated to compositions of morphisms, and in particular $\epsilon$-maps.  

\small
\begin{eqnarray*}
\ov{\mbox{Mary snores}} &=& \F(n \cdot n^r \cdot  s)(\ov{\textmd{Mary}} \otimes \ol{\textmd{snores}}) \\
&=&(\epsilon^r_N \otimes 1_S)(\ov{\textmd{Mary}} \otimes \ol{\textmd{snores}}) 
\end{eqnarray*}
\normalsize

%
A simple computation shows that the above is equal to $\ov{\textmd{Mary}} \times \ol{\textmd{snores}}$; similarly, for  the meaning of a transitive sentence we obtain:

\begin{equation*}
\ov{\mbox{Mary likes musicals}} = \ov{\textmd{Mary}} \times \ol{\textmd{likes}} \times \ov{\textmd{musicals}}
\end{equation*}

Note that tensor contraction (in spaces with fixed basis) is associative, so there is no need to keep track of brackets in the above. The situation  is similar to pregroups, where the monoid multiplication is again associative. 

\subsection{Frobenius algebras}
\label{sec:frob}

Compact closed categories on their own do not have much structure: there is a binary operation and the maps $\epsilon$ and $\eta$. The expressive power of these categories can be increased using Frobenius algebras. We define these below. 


Given a compact closed category ${\cal C}$, an object $X \in {\cal C}$ has a Frobenius structure on it if there exist  the following morphisms:

\vspace{-0.6cm}
\begin{align*}
\Delta \colon X \to X \otimes X&\qquad& \iota \colon X \to I\\
\mu \colon  X \otimes X \to X  &\qquad& \zeta \colon I \to X
\end{align*}
\vspace{-0.6cm}

These have to satisfy certain conditions, the most important to us being the \emph{Frobenius condition}:

\vspace{-0.6cm}
\begin{align*}
\mbox{\footnotesize $(\mu \otimes 1_X) \circ (1_X \otimes \Delta) \ = \  \Delta \circ \mu  \ = \  (1_X \otimes \mu) \circ (\Delta \otimes 1_X)$}
\end{align*}
\vspace{-0.6cm}

Vector spaces with fixed basis do have such structures over them, generally referred to by \emph{copying} and \emph{merging}.  For $\ov{v} \in V, \ol{w} \in V \otimes V$, we have that $\Delta(\ov{v})\in V \otimes V$ is a diagonal matrix whose diagonal elements are weights of $\ov{v}$, and $\mu(\ol{w}) \in V$ is a vector
consisting only of the diagonal elements of $\ol{w}$.


These structures have been used in previous work to encode lower dimensional verb matrices into higher dimensional tensors \cite{kartsaklis2012,kartsaklis2014} and to pass the information around sentences with relative clauses by copying and merging \cite{relpronouns1,relpronouns2}.

\subsection{Graphical calculus}
\label{sec:calculus}

In the presence of higher order tensor product spaces, calculations can become quite complex. The formalism of compact closed categories and Frobenius structures is complete with regard to a graphical calculus 
\cite{selinger2011survey} that simplifies the computations to a great extend. We briefly overview the main components of this language. 

Objects are depicted by lines and morphisms  by boxes.  Tensor products between objects and morphisms are given by
juxtaposition of their diagrams, while composition of morphisms amounts to connecting outputs to inputs.  Examples are as follows:

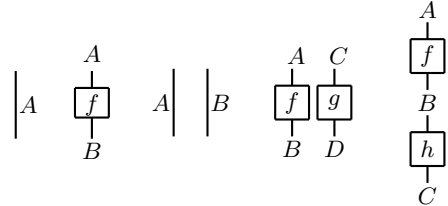
\footnotesize
\begin{center}
  
\InputIfFileExists{compact-diag.tikz}{}{\input{.//tikz//compact-diag.tikz}}
\quad
  
\InputIfFileExists{compact-diag-tensor.tikz}{}{\input{.//tikz//compact-diag-tensor.tikz}}

\end{center}
\normalsize

The $\epsilon$ maps are depicted by cups, $\eta$ maps by caps, and yanking by their composition and straightening of the strings. For instance:

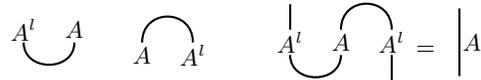
\footnotesize
\begin{center}
  
\InputIfFileExists{compact-cap-cup.tikz}{}{\input{.//tikz//compact-cap-cup.tikz}}

    \qquad
    
\InputIfFileExists{compact-yank.tikz}{}{\input{.//tikz//compact-yank.tikz}}

\end{center}
\normalsize

The diagrams corresponding to the Frobenius  morphisms are as follows:

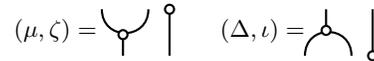
\footnotesize
\ctikzfig{comp-alg-coalg}
\normalsize
 
\noindent
with the Frobenius condition being depicted as:

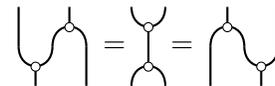
\ctikzfig{frobcond}

The defining axioms guarantee that any picture of a Frobenius computation can be reduced to a normal form (so-called a ``spider'') that only depends on the number of input and output strings of the nodes: 

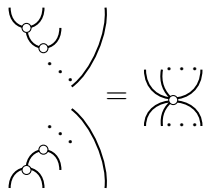
\[

\InputIfFileExists{spider.tikz}{}{\input{.//tikz//spider.tikz}}

\]

Elements within the objects (for the case of vector spaces, vectors) are depicted  by  morphisms from the unit. These are shown by triangles with a number of strings emanating from them. The number of strings denotes the order of the tensor; for instance, the diagrams for $\ov{v} \in V, \ol{v'} \in V \otimes W$, and $\ol{v''} \in V \otimes W \otimes Z$ are as follows:

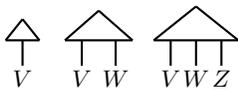
\footnotesize
\ctikzfig{compact-diag-triangle}  
\normalsize
%
%
%

\section{Information structure and intonation}
\label{sec:infstruct}

The term \textit{information structure} collectively refers to techniques that aim to enhance the communication between two interlocutors in order to optimize the conveyed message for the benefit of the addressee \cite{chafe}. One such technique, for example, is to emphasize a particular part of the utterance that is important for the listener by changing the spoken pitch:

\begin{exe}
  \ex Q: What does Mary like? \\
      A: Mary likes {\sc musicals}
  \label{ex:fel}      
\end{exe}

The emphasis imposes a specific information structure to the uttered sentence, essentially splitting it in two parts: The part in upper-case above is what \newcite{steedman2000information} calls \textit{rheme}---the information that the speaker wishes to make common ground for the listener; the rest of the sentence, i.e. what the listener already knows, is called \textit{theme}. The question in (\ref{ex:fel}) puts the listener in a specific attentional state, in the context of which an answer such as:

\begin{exe}
  \ex A: \#{\sc Mary} likes musicals
\end{exe}

\noindent
will be \textit{infelicitous}, that is, not compatible with that state.

The distinction between theme (or topic) and rheme (or comment) has great significance from an information structure point view, since it defines a generic shape for the sentence that directly reflects the attentional needs of the addressee. A further dimension that can be found in both rheme and theme distinguishes between the \textit{focus}, that is, the specific word that receives most of the intonational emphasis, and the background, which consists of the rest of the words in the specific text segment. Note that in contrast to rheme/theme distinction, focus and background seem to operate at the lexical level.\footnote{Actually many authors use the term \textit{focus} as a synonym for \textit{rheme}; the definitions we give in this paper follow \cite{steedman2000information}.}

Furthermore, we should point out that although the examples we use in this paper are mainly based on question/answer dialogues, this is not by any means the only case where the presence of a specific information structure can be useful. For example, consider the dialogue:

\begin{exe}
  \ex --I think Mary likes jazz. \\
      --Mary likes {\sc musicals}.
\end{exe}

%

Information structure can be expressed in different ways that may vary from language to language. In English, for example, the means for defining information structure is \textit{intonation}: variations of spoken pitch, the purpose of which is to emphasize parts of the utterance that might be important for the conveyed message, as we saw above in our examples. However, in other languages such as Japanese or Cantonese, the intonational boundaries can be also specifically marked by morphological devices, e.g. special particles \cite{fery2008}. Finally, the position of a text segment in a sentence can also be an indication of its information-structural role. In English, for example, themes tend to appear at the beginning of a clause. 


In this paper we concentrate on the sentence-level distinction between rheme and theme. 

\section{Grammar and intonation}

The presence of a distinct layer of information structure that seems to co-exist with the grammatical structure of a sentence, poses the interesting question regarding the exact relationship that holds between those different structural aspects. For example, although the text segment  ``Mary likes'' forms a perfectly acceptable theme, most linguists would agree that it does not also comprise a valid grammatical constituent. In spite of this claim, though, it is interesting to note that a number of categorial grammars, including Combinatory Categorial Grammar (CCG) \cite{steedman}, treat text segments like the above as possible syntactic constituents. Consider the following ditransitive sentence:

\begin{exe}
\ex\label{ex:john} John gave Mary a flower
\end{exe}

In CCG, this sentence has a number of different syntactic derivations, two of them are the following:

\footnotesize
\begin{equation}
\begin{minipage}{0.80\linewidth}
\begin{center}  
\deriv{4}{
{\rm John} & {\rm gave} & {\rm Mary} & {\rm a~flower} \\
\uline{1} & \uline{1} &\uline{1} & \uline{1} \\
\textsc{np} & \textsc{((s\bs np)/np)/np} & \textsc{np} & \textsc{np} \\
   & \fapply{2} & \\
   & \cmc{2}{\textsc{(s\bs np)/np}} & \\
   & \fapply{3} \\
   & \cmc{3}{\textsc{s\bs np}} \\
\bapply{4} \\
\cmc{4}{\textsc{s}} \\
}  
\end{center}
\end{minipage}
\label{fig:spur1}
\end{equation}
\normalsize

\footnotesize
\begin{equation}
\begin{minipage}{0.80\linewidth}
\begin{center}  
\deriv{4}{
{\rm John} & {\rm gave} & {\rm Mary} & {\rm a~flower} \\
\uline{1} & \uline{1} &\uline{1} & \uline{1} \\
\textsc{np} & \textsc{((s\bs np)/np)/np} & \textsc{np} & \textsc{np} \\
\ftype{1} & \fapply{2} & \\
{\textsc{s/(s\bs np)}} & \cmc{2}{\textsc{(s\bs np)/np}} & \\
\fcomp{3} & \\
\cmc{3}{\textsc{s/np}} & \\
\fapply{4} \\
\cmc{4}{\textsc{s}} \\
}  
\end{center}
\end{minipage}
\label{fig:spur2}
\end{equation}
\normalsize

Note that (\ref{fig:spur1}) proceeds by first composing the part corresponding to the verb phrase (``gave Mary a flower''); later, in the final step, the verb phrase is composed with the subject `John'. The situation is reversed for (\ref{fig:spur2}), where the use of type-raising and composition rules of CCG allow the construction of the fragment ``John gave Mary'' as valid grammatical text constituent, which is later combined with the direct object of the sentence (`a flower'). \newcite{steedman2000information} argues that this form of different syntactic derivations that one can get even for very simple sentences when using CCG (some times referred to with the somewhat belittling term ``spurious readings''), actually serve to reflect variations in  information structure. Each one of the above derivations subsumes a different intonational pattern, distinguishing the rheme from the theme when the sentence is used for answering different questions:  (\ref{fig:spur1}) answers to ``Who gave Mary a flower?'', whereas (\ref{fig:spur2}) to ``What did John give to Mary?''. 

In other words, the claim here is that (a) surface structure and information structure coincide; and (b) the role of information structure is to provide a particular interpretation of the surface structure. Let us define this important idea in a precise way, since it will be the cornerstone of the model presented in this paper:

\begin{postulate}
Intonational boundaries in an utterance determine the intended syntactic structure.
\label{pst:axiom}
\end{postulate}

In our grammatical formalism, pregroup grammars, variations in a grammatical derivation similar to above are only implicitly assumed, since the order of composition remains unspecified. This fact is apparent in the pregroup derivation of the example sentence, where both (\ref{fig:spur1}) and (\ref{fig:spur2}) are subsumed into the following \textit{reduction diagram}:

\begin{equation}
   \footnotesize
   
\InputIfFileExists{dtrans-preg.tikz}{}{\input{.//tikz//dtrans-preg.tikz}}

   \normalsize
\end{equation}

Furthermore, it is directly reflected in our semantic space through the functorial passage, via the fact that tensor contraction is associative:

\vspace{-0.4cm}
\small
\begin{eqnarray}
 \ov{\textmd{John}} \times  ( \ol{\textmd{gave}} \times \ov{\textmd{Mary}} \times \ov{\textmd{flower}} ) & = \label{equ:transparency} \\
  ( \ov{\textmd{John}} \times \ol{\textmd{gave}} \times \ov{\textmd{Mary}} ) \times \ov{\textmd{flower}} \nonumber 
\end{eqnarray}
\normalsize
\vspace{-0.4cm}

Eq. \ref{equ:transparency} constitutes a natural manifestation of the \textit{principle of combinatory transparency} \cite{steedman}: no matter in what order the various text constituents are combined, the semantic representation assigned to the sentence is always the same; in other words, information structure should not affect semantic conditions. Note, however, that even in the strict setting of formal semantics this is not always the case. Consider the behaviour of the following sentence under the presence of the focus-sensitive particle `only':

\begin{exe}
  \ex \begin{xlist}
          \ex\label{ex:john1} John only gave Mary {\sc a flower}
          \ex\label{ex:john2} John only gave {\sc Mary} a flower
       \end{xlist}   
\end{exe}

The use of different intonational focus clearly changes the semantic value of the sentence: (\ref{ex:john1}) is true if the only thing that John gave to Mary was a flower (but he might have given things to other girls as well), while (\ref{ex:john2}) is true if the only person who got a flower from John was Mary. 

In the more relaxed and quantitative setting of a compositional distributional model of meaning, the idea of having vectorial representations of words and sentences that reflect intonational patterns seems even more legitimate. This concept is aligned with the distributional nature of such models: given a text corpus containing information structure annotations (of any kind), we would assume that the co-occurrence vector of a word under focus (say, $\ov{\textsc{book}}$) would slightly differ from that of the vector representing the normal use of the word ($\ov{\textmd{book}}$).\footnote{In the trivial case, this would be true by the presence of intonational markers in the immediate context of a word under focus, as opposed to its normal use.}
Furthermore, we would expect that, after the composition, this difference would be also reflected in the vector representing the meaning of the entire sentence. From the next section we start working towards imposing this behaviour on the categorical model of \newcite{Coeckeetal}. 

\section{Intonation in pregroups}

Traditionally, a notational system describing intonation consists of markings that indicate pitch accents and boundaries. Using the notation of \newcite{pierrehumbert1990}, for example, we get the following for our example sentence:\footnote{Example taken from \cite{steedman2000information}.}\footnote{From now on we explicitly mark themes and rhemes in our examples for clarity.}

\begin{exe}
  \ex \rheme{{\sc Mary}}~~~\theme{likes {\sc musicals}} \\
          $~~~~$H^*~L $~~~~~~~~~~~~~~~~~$L+H^*~LH\%
\end{exe}

The prosody starts with a sharp pitch accent (H^*) that puts the focus on `Mary', and continues with a rapid fall to low pitch (L boundary) that signifies a transition from rheme to theme. Within theme now, the focus goes to `musicals' which gets the less rapidly rising pitch L+H^*, whereas the boundary LH\% expresses a rising continuation that marks the end of theme. In the case that theme precedes the rheme, we have the following pattern:

\begin{exe}
  \ex\label{ex:mary}  \theme{{\sc Mary} likes}~~~\rheme{{\sc musicals}} \\
                         $~~~~$L+H* LH\% $~~~~~~~~~~~$ H^* LL\%
\end{exe}

As mentioned earlier, this paper mainly addresses the rheme/theme aspect of information structure, which is directly related to boundary markings. We start by representing an intonational boundary using a special token $\triangleright$, for which the following relations hold:

\begin{equation}
\label{boundaries}
  \textit{theme}~\triangleright~\textit{rheme}~~~~~\text{or}~~~~~\textit{rheme}~\triangleleft~\textit{theme}
\end{equation}

Naturally, $\triangleright$ is equivalent to LH\%, while $\triangleleft$ corresponds to L in the \newcite{pierrehumbert1990} notation. It is very important to emphasize at this point that the above introduced tokens are far from an ad-hoc means for achieving a goal. Recall from our discussion in Section \ref{sec:infstruct} that while in English the means of imposing information structure is purely phonological, this is not necessarily the case for other languages. As a concrete case, in Buli (a Gur language spoken in Ghana), the rheme is preceded by a \textit{focus marker}, which again can be interpreted as an information-structural boundary since it separates the theme from rheme; this is shown in the following example \cite{fiedler2006}:

\begin{exe}
  \ex Q: What did the woman eat? \\
      A: \`\textopeno~~~~~~\textipa{\ng}\`\textopeno b~~~\textbf{k\`a}~~~~~~~\textbf{t\'u\'e} \\
      $~~~~~~$3sg$~~$eat$~~~$(FM)$~~~$ beans
\end{exe}

To formalize this mixing of syntactical and information structure in the context of a pregroup grammar, we add the two boundary markers  to the vocabulary and introduce two new atomic types:

\vspace{-0.3cm}
\begin{equation}
   \text{Theme:~} \theta  \qquad \text{Rheme:~} \rho 
\end{equation}
\vspace{-0.3cm}

An intonation pregroup grammar then will have the following form:

\[
P(\Sigma \cup \{\triangleright,\triangleleft\}, \{n,s, \theta,\rho\})
\]

For the case of a simple transitive sentence, we get the following boundary types, based on the fact that now the  boundary (and not the verb) becomes the head of our sentence:

\vspace{-0.3cm}
\begin{equation}
  \triangleright: \theta^r \cdot s \cdot  \rho^l \qquad \triangleleft: \rho^r \cdot s \cdot  \theta^l
\end{equation}
\vspace{-0.3cm}

The type dictionary changes accordingly:  a transitive verb such as  `like' will be assigned two more types $n^r \cdot \theta$ and $\theta \cdot n^l$ depending whether it produces a left-hand theme or a right-hand theme in the sentence; similarly, nouns will be assigned  the extra type  $\rho$. For the two cases of Eq. \ref{boundaries}, we  obtain the following derivations:

\begin{equation}
  \footnotesize
  
\InputIfFileExists{inton-preg.tikz}{}{\input{.//tikz//inton-preg.tikz}}

  \normalsize
  \label{equ:into-preg}
\end{equation}

\begin{equation}
  \footnotesize
  
\InputIfFileExists{inton-preg2.tikz}{}{\input{.//tikz//inton-preg2.tikz}}

  \normalsize
  \label{equ:into-preg2}
\end{equation}

After transferring this to $\mathbf{FVect}$ via our functor in Eq. \ref{equ:functor}, and extending its action on atomic types by defining $\F(\theta) = \Theta$ and $\F(\rho) = P$,  we get the obvious semantic counterpart:

\begin{equation}
  \footnotesize
  
\InputIfFileExists{inton-fvect.tikz}{}{\input{.//tikz//inton-fvect.tikz}}

  \normalsize
  \label{equ:into-fvect}
\end{equation}

There are some important observations based on the derivation in (\ref{equ:into-fvect}) above. Firstly, our simple sentence now is given in terms of a theme and a rheme, as required, both of which contribute equally to its construction. Additionally, note that our verb is not any more a function of two arguments (of a subject and an object) as in the canonical case, but of a single noun: it takes as input a subject in order to return a theme. Hence, in contrast to a typical case of a transitive verb, the semantic representation of which requires a tensor of order 3, in this case the corresponding linear map takes the form $N \to \Theta$, which can be canonically represented by a \textit{matrix} $N\ten \Theta$.

The question of how to properly model intonation in compositional distributional semantics is evidently epitomized in choosing an appropriate form for the tensor of the $\triangleright$ token in (\ref{equ:into-fvect}). In order to provide an answer to this, we first need to examine the concepts of rheme and theme from a semantic point of view. 

\section{A semantic  truth-theoretic argument}
\label{sec:set-example}

We use as an example the following simple case:

\begin{exe}
  \ex Q: Who does John like? \\ 
      A: \theme{John likes} \rheme {\sc Mary}
\end{exe}

From an extensional point of view, the semantic value of the theme can be seen as a set of \textit{alternative} options \cite{rooth1992}, each one of which may be used as a response to the given question:

\vspace{-0.5cm}
\begin{equation*}
  \sem{\text{John (might) like}} = \{x|\text{John (might) like}~x\}
\end{equation*}
\vspace{-0.5cm}

As a consequence, the role of the rheme now is to \textit{restrict} the set of alternatives to a specific choice \cite{steedman2000information}. Note that this action of restricting the available choices is responsibility of the intonational boundary; indeed, the boundary can be seen as a binary operator that performs the merging of the theme with the rheme, restricting the alternatives set of the theme to a specific response:

\begin{exe}
   \ex \theme{John likes} $\triangleright$ \rheme {{\sc Mary}} := \\
   $\triangleright($\theme{John likes},\rheme{\sc Mary}$)$
\end{exe}

This is  what Diagram (\ref{equ:into-fvect}) shows; in our multi-linear setting, the boundary becomes a bi-linear map $\Theta\otimes P \to S$ that performs the required ``restriction''. Now, what is the most appropriate way to model this operation in the extensional setting discussed above? Note that by simply checking if rheme is contained in the alternatives set is not sufficient; this would return \textit{true} or \textit{false} as an answer to a question that expects a person. A more appropriate choice then is to model the boundary by using set intersection: we take the meaning of rheme to be a singleton that contains the answer, and the meaning of the sentence to be the intersection of rheme with theme:

\vspace{-0.2cm}
\begin{equation}
  \label{equ:intersection}
  \{\text{Mary}\} \cap \{x|\text{John (might) like}~x\}
\end{equation}

The answer will be again the singleton $\{\text{Mary}\}$ if Mary is included in the set of people who John potentially likes, and the empty set otherwise. Thus we have achieved our goal: the theme set has been restricted according to the provided response. We generalize this argument to an arbitrary pair of rheme and theme (with $S_{\text{theme}}$ denoting theme's corresponding alternative set) as follows:

\vspace{-0.5cm}
\begin{equation}
\begin{aligned}
  \triangleright(\text{rheme},\text{theme}) = \{\text{rheme}\} \cap S_{\text{theme}} \\
  \left\{ \begin{array}{lr} \text{rheme} & \text{if~rheme} \in S_{\text{theme}} \\
  \varnothing & o.w. \end{array} \right . 
\end{aligned}
\end{equation}

\subsection{From sets to vector spaces}

We transfer the above reasoning to vector spaces, by encoding sets and relations in vectorial forms. The vectorial form of a set is a vector space (let it be $N = \{n_i\}_i$) whose basis vectors are the elements of the set.  For the sake of demonstration (and this will become clear as the section reads on), we define our sentence space to be   a one dimensional space where the origin denotes falsity and everything else denotes truth. One can take this to be a dimension in any vector space; here we take it to be in $N$  and  denote its  basis vector with a basis vector of $N$. Furthermore, a binary relation such as $likes(x,y)$ can be represented as an adjacency matrix $\textbf{W}$ in which $W_{ij}$ is 1 if the pair $(i,j)$ is contained in the relation and 0 otherwise. Note that this matrix is isomorphic to a tensor in $N\ten S \ten N$, since our sentence space is one-dimensional.

Let us apply categorical composition to compute a vectorial representation for the theme of our sentence, ``John likes''.

\vspace{-0.3cm}
\small
\begin{eqnarray}
  (\epsilon^r_N \ten 1_S) \left( \ov{n_3} \ten \left( \sum\limits_{ij} W_{ij} \ov{n_i}\ten \ov{n_j} \right)  \right) & = \nonumber \\
  \sum\limits_{ij} W_{ij} \langle \ov{n_3}|\ov{n_i} \rangle \ov{n_j} =  
  \sum\limits_{ij} W_{ij} \delta_{3i} \ov{n_j} = \sum\limits_{j} W_{3j} \ov{n_j}  \label{equ:cat-theme}&
\end{eqnarray}
\normalsize

\vspace{-0.1cm}

Hence the vectorial representation of ``John likes'' becomes indeed the subset of all individuals who might be liked by the person denoted by vector $\ov{n_3}$, and can be seen as the semantic value of the theme of our sentence. The next step is to compose this theme with the rheme `Mary'; in other words, we must decide an appropriate type of composition for our intonational boundary. Let us first try again standard categorical composition:

\vspace{-0.4cm}
\small
\begin{eqnarray}
  (1_S\ten\epsilon_N^l) \left( \left( \sum\limits_{j} W_{3j} \ov{n_j} \right) \ten \ov{n_1} \right) & = \nonumber \\ 
  \sum_j W_{3j} \langle n_j|n_1 \rangle = \sum_j W_{3j} \delta_{j1} = W_{31}
\end{eqnarray}
\normalsize
\vspace{-0.2cm}

Note that this corresponds to a set membership test; the result is 1 if Mary is included in the set of alternative responses and 0  otherwise. However, as noted before, in information structure terms a more appropriate operation would be to take the intersection of the singleton $\{\text{Mary}\}$ with the set of alternatives. Interestingly, set intersection now corresponds to element-wise vector multiplication (in this work denoted by symbol $\odot$) and the vector space equivalent of Eq. \ref{equ:intersection} becomes:

\vspace{-0.2cm}
\small
\begin{equation}
   \left( \sum\limits_{j} W_{3j} \ov{n_j} \right) \odot \ov{n_1} = 
   \left\{ \begin{array}{lr} \ov{n_1} & \text{if~}W_{31}=1 \\ \ov{0} & o.w. \end{array} \right.
\end{equation}
\normalsize
\vspace{-0.2cm}

The result is now `Mary', if Mary is included in the set of valid answers, and the zero vector otherwise. The fact that the meaning of our sentence becomes an element of the noun space demonstrates clearly that, in information structure terms, there is a necessity for a shared vector space between sentences and nouns (or noun phrases)---a direct consequence of the fact that now the meaning of a sentence is mainly focused on a specific noun or noun phrase therein. Furthermore, since a sentence is now expressed as a merging of a theme and a rheme, it is also required that $\Theta=S=P$ (and equal to what we took to be $N$ in the preceding). In the next section we encode the above reasoning in the abstract form of compact closed categories and then present an instantiation in vector spaces. 

\section{Intonation in compact closed categories with Frobenius structure}

The point with regard to shared spaces is accomplished by  the following types assignment:

\begin{equation}
  \mathcal{F}(x) = W \qquad \forall x \in \{n,s,\theta,\rho\}
\end{equation}

As a consequence of the above, the vector spaces assigned to transitive verbs are computed as follows:

~
\vspace{-0.4cm}
\begin{equation*}
\F(n^r\cdot \theta) = \F(\theta \cdot n^l) = W \otimes W
\end{equation*}

Furthermore, boundaries are assigned to the following vector space:

\vspace{-0.5cm}
\begin{equation*}
\F(\theta^r\cdot s\cdot \rho^l) = \F(\rho^r\cdot s\cdot \theta^l) = W \otimes W \otimes W
\end{equation*}

We have now arrived at a central point of this paper. As the semantic representation of a boundary, we assign the following morphism:

\begin{equation}
(1_W \ten \mu_W \ten 1_W) \circ (\eta_W \ten \eta_W)
\label{equ:boundary-morph}
\end{equation}

Note that the above is indeed  an element in $W \otimes W \otimes W$:

\vspace{-0.5cm}
\begin{eqnarray}
  \triangleright,   \triangleleft: I \cong I\ten I \xrightarrow{\eta_W\ten\eta_W} W\ten W \ten W \ten W \\
  \xrightarrow{1_W\ten\mu_W\ten 1_W} W\ten W\ten W \nonumber
\end{eqnarray}

The reasoning behind our assignment will become clear in a moment. For now, we proceed to a formal definition:

\begin{definition}
\label{manindef}
  The meaning vector of a sentence expressed in information structure terms is given by:
  
  \vspace{-0.3cm}
  \footnotesize
  \begin{equation}
     (1_W \ten \mu_W \ten 1_W) \circ (\eta_W \ten \eta_W)(\ov{\textmd{theme}}\ten \ov{\textmd{rheme}})
  \end{equation}
  \normalsize
  
  \noindent
  when theme precedes the rheme, or  as follows in the opposite case.

  \vspace{-0.3cm}  
  \footnotesize
  \begin{equation}
     (1_W \ten \mu_W \ten 1_W) \circ (\eta_W \ten \eta_W)(\ov{\textmd{rheme}}\ten \ov{\textmd{theme}})
  \end{equation}
  \normalsize  
\end{definition}

\noindent
These vectors are depicted as follows:  

\vspace{-0.4cm}
\begin{equation}
  \footnotesize
  
\InputIfFileExists{boundary1.tikz}{}{\input{.//tikz//boundary1.tikz}}

  \normalsize
  \label{equ:frob-is1}
\end{equation}
\vspace{-0.3cm}
\begin{equation}
  \footnotesize
  
\InputIfFileExists{boundary2.tikz}{}{\input{.//tikz//boundary2.tikz}}

  \normalsize
  \label{equ:frob-is2}
\end{equation}

Note that the normal forms at the right-hand side of the diagrams above are  direct applications of the Frobenius condition. Furthermore, either the theme or rheme here might correspond to large text constituents, i.e. phrases or even sentences. In this case, the proposed framework guarantees that an appropriate vector will be created for them based on categorical composition.  

\section{Vector space instantiation}

Our justification for using the semantic form of Eq. \ref{equ:boundary-morph} for the boundary comes from the fact that it produces normal forms as below:
\begin{equation}
  \mu(\ov{\textmd{theme}}\ten \ov{\textmd{rheme}})~~~\mu(\ov{\textmd{rheme}}\ten \ov{\textmd{theme}})
\end{equation}

This is exactly how element-wise vector multiplication is defined from a categorical perspective:

\begin{equation}
  
\InputIfFileExists{frob-mu.tikz}{}{\input{.//tikz//frob-mu.tikz}}

\end{equation}

As a result, the linear algebraic instantiations of Definition \ref{manindef} become as follows:
\[
\overrightarrow{\textmd{rheme}} \odot \overrightarrow{\textmd{theme}}\qquad
\overrightarrow{\textmd{theme}} \odot \overrightarrow{\textmd{rheme}}
\]

We stress again the fact that rheme  and theme can have complex structures, and their vector meanings will reflect this strutter.  For  simple transitive sentences\footnote{We provide more complicated examples later in Sect. \ref{sec:complex}.} of the form ``subject verb $\triangleright$ object'' or ``subject $\triangleleft$ verb object'', we get linear algebraic meanings as follows:
\[
{\small
(\ov{\textmd{subj}} \times \ol{\textmd{verb}}) \odot \ov{\textmd{obj}}\quad\quad
\ov{\textmd{subj}} \odot  (\ol{\textmd{verb}} \times \ov{\textmd{obj}})
}\]

\noindent
As an example of a composed theme, consider:

\vspace{-0.4cm}
\begin{equation}
  \footnotesize
  
\InputIfFileExists{into-der1.tikz}{}{\input{.//tikz//into-der1.tikz}}

  \normalsize
  \label{equ:frob-bound}
\end{equation}
\vspace{-0.4cm}

A vector is computed for the theme `Mary likes' according to the rules of the grammar, and then this vector is element-wise multiplied with the vector of the rheme (which, in this example, is just the distributional vector of the word). 

\section{Interpretation}

The transition from the set-theoretical framework to high dimensional real vector spaces poses the question what is the role of  element-wise vector multiplication in the latter setting. Compositional models based on element-wise vector addition or multiplication are usually referred to as \textit{vector mixture} models---a term that emphasizes on the \textit{equal contribution} of each word to the final result, which produces a kind of average of the input vectors. Note that this behaviour stands in direct contrast with the categorical compositional approach, in which the type-logical identities of words strictly depend on their grammatical role. Due to their simplicity, vector mixture models have been studied extensively \cite{Lapata}, demonstrating steady and reasonably good performance in a number of tasks. 

The significance of the Frobenius operators for our model (as opposed to some other form of combinatory mechanism) is that their concrete manifestation in a vector space setting imposes exactly this vector mixture behaviour, in the form of element-wise vector multiplication. In other words, the result is a combination of two compositional approaches, vector mixtures and categorical models, in a unified framework: while categorical composition is still applied to compute vectorial representations for a theme and a rheme, the two parts contribute equally to the final result via element-wise multiplication imposed by the Frobenius operators. This puts the necessary focus on the appropriate part of the sentence, reflecting the variation in meaning intended by the intonational pattern.


To what extent the notion of a rheme as a means for restricting the theme applies in $\mathbf{FVect}$? Note that, from a geometric perspective, element-wise vector multiplication acts as a scaling of the basis; for example, $\left( \begin{smallmatrix} x \\ y \end{smallmatrix} \right) \odot \left( \begin{smallmatrix} 2.0 \\ 0.5 \end{smallmatrix} \right)$ transforms the vector space in which the first vector lives so that the units on the $x$-axis are doubled and the units on the $y$-axis are halved.\footnote{Of course we can think of a similar scaling taking place on the two axes of the second vector by factors $x$ and $y$.} Furthermore, a zero value in one vector would completely eliminate the corresponding component in the other. Hence, the concept of restricting the theme has now taken a new quantitative form, generalizing appropriately our initial intuition (motivated by set intersection) to the multi-dimensional, real-valued setting of $\mathbf{FVect}$.

\section{Relation to previous work}

How does the above derivations correlate to the premises of the original framework, in which `likes' is a transitive verb with type $n^r \cdot s \cdot n^l$? Note that another application of the Frobenius condition on the normal form of Diagram (\ref{equ:frob-bound}) will give us:

~
\vspace{-0.6cm}
\begin{equation}
  \footnotesize
  
\InputIfFileExists{into-der2a.tikz}{}{\input{.//tikz//into-der2a.tikz}}

  \normalsize
  \label{equ:copy-obj}
\end{equation}
\vspace{-0.4cm}

In other words, the semantic representation of word `likes' can be still regarded as a bi-linear map, faithfully encoded in a tensor of order 3, as required by the framework. In this case, the tensor of `likes' in $\mathbf{FVect}$ is seen as created by applying the morphism $1_W \ten \Delta_W$ on a matrix representing the verb `likes'. The limitation, of course, is that now the middle wire carrying the result (the sentence vector space) cannot be any more differentiated from the two argument wires (the noun vector spaces), since it is produced by copying one of them.

Note that these are the Frobenius models of \newcite{kartsaklis2014}, referred to as Copy-Subject and Copy-Object, and originally used as a means for faithfully encoding a verb matrix to a tensor of order 3, thus restoring the functorial relation between the semantic representation and the grammatical type. The present theory\footnote{An early account  of which also appears in the doctoral thesis of the first author \cite{kartsaklisphd}.} offers an alternative more complete account that goes far beyond providing a convenient way to expand a matrix to a cube.

\section{Covering complex intonational patterns}
\label{sec:complex}

So far we examined simple cases of intonation, in which our sentence consisted of a single rheme and a theme. In this section we turn our attention to some more interesting examples.


\subsection{Multiple rhemes}

We will first examine the case of a sentence with more than one rhemes. Imagine the following question/answer dialogue:

\begin{exe}
  \ex\label{ex:tworhemes} Q: Who likes whom? \\
      A: \rheme{{\sc John}} \theme{likes} \rheme{{\sc Mary}}
\end{exe}

In our pregroup notation, this introduces two distinct  intonational boundaries in the sentence. The derivation takes the following form:

\vspace{-0.2cm}
\begin{equation}
  \footnotesize
  
\InputIfFileExists{tworhemes-preg.tikz}{}{\input{.//tikz//tworhemes-preg.tikz}}

  \normalsize
  \label{equ:tworhemes-preg}
\end{equation}
\vspace{-0.2cm}

Note that the type of `likes' now becomes $\theta\cdot\theta$; in other words, the theme is not any more a function (no adjoint is present in the type), but a \textit{higher order} atomic entity. This is directly reflected in $\mathbf{FVect}$ where we get:

\vspace{-0.4cm}
\begin{equation}
  \footnotesize
  
\InputIfFileExists{tworhemes1.tikz}{}{\input{.//tikz//tworhemes1.tikz}}

  \normalsize
  \label{equ:tworhemes}
\end{equation}
\vspace{-0.4cm}

The result of this computation is now a matrix and not a vector. Indeed, if we follow the linear algebraic calculations we get:

\vspace{-0.5cm}
\begin{eqnarray}
   (\mu_W\ten \mu_W)(\ov{\textmd{John}}\ten \ol{\textmd{likes}}\ten \ov{\textmd{mary}}) & = \\
   (\ov{\textmd{John}}\ten\ov{\textmd{Mary}})\odot \ol{\textmd{likes}} \nonumber
\end{eqnarray}
\vspace{-0.5cm}

The behaviour above follows the premises of the proposed model: Since our theme is a matrix, the calculations follow naturally, producing another matrix as the rheme (the tensor product of the two individual rhemes) that \textit{restricts} as required the theme via element-wise multiplication.
Note that this means that a sentence with one rheme would not be comparable with a sentence with two rhemes, since it would live in a different space. That is again not surprising: the shape of theme defines the shape of the sentence vector space, and only themes of the same order can be compared to each other.

\subsection{Relational words as rhemes}

We have conveniently avoided to discuss until now the case in which the rheme is not a noun phrase, but a relational word as below:

\begin{exe}
  \ex Q: How does John feel about Mary? \\
      A: \theme{John} \rheme{{\sc likes}} \theme{Mary}
\end{exe}

In pregroups we model such a situation by the following derivation:

\begin{equation}
  \footnotesize
  
\InputIfFileExists{rheme-verb.tikz}{}{\input{.//tikz//rheme-verb.tikz}}

  \normalsize
\end{equation}

Note that this time the verb becomes a higher order rheme, getting the type $\rho\cdot\rho$. However, when this is transferred to $\mathbf{FVect}$ the symmetry of the category and the commutativity of the Frobenius algebra means that the vector of the sentence becomes equal to that of Example (\ref{ex:tworhemes}). In general, problems due to commutativity of the Frobenius operators can be resolved if one moves to non-commutative versions of Frobenius algebras. \newcite{piedeleu2015open-calco} explore such constructions in the context of language by elevating the categorical model of \newcite{Coeckeetal} to an open quantum system setting, in which words are represented as mixed states.

\subsection{Nested rhemes}

Consider the following case:

\begin{exe}
  \ex What was the book Mary wrote about? \\
     \theme{Mary wrote a book about} \rheme{{\sc art}}
\end{exe}

The interesting point here is that the intonational boundary is placed in a position that constitutes a glaring violation of the grammatical structure, which in the normal case has the following form:

\begin{equation}
  \footnotesize
  
\InputIfFileExists{nested-normal.tikz}{}{\input{.//tikz//nested-normal.tikz}}

  \normalsize
  \label{equ:nested-normal}
\end{equation}

For cases like these we should recall that our framework is entirely built on the assumption of Postulate \ref{pst:axiom}: in the context of information structure, intonational boundaries \textit{determine} the intended syntactical structure. For our case, we get:

\begin{equation}
  \footnotesize
  
\InputIfFileExists{nested.tikz}{}{\input{.//tikz//nested.tikz}}

  \normalsize
  \label{equ:nested}
\end{equation}

The linear-algebraic result follows trivially as the usual element-wise composition of the theme with the rheme.

\subsection{Rheme in the middle of sentence}

In many cases a noun phrase can serve as the rheme while being placed in the middle of the sentence, splitting the theme into two parts:

\begin{exe}
  \ex Did Mary write an essay about art? \\
     \theme{Mary wrote} \rheme{{\sc a book}} \theme{about art}
\end{exe}

In these cases, the left-hand intonational boundary gets the type $\theta^r \cdot \rho \cdot \rho^l$, as below:
\begin{equation}
  \footnotesize
  
\InputIfFileExists{nested1.tikz}{}{\input{.//tikz//nested1.tikz}}

  \normalsize
  \label{equ:nested1}
\end{equation}

In other words, a new rheme is produced that is used as input to the right-hand intonational boundary. In $\mathbf{Fvect}$ we get the following interaction:

~
\vspace{-0.3cm}
\begin{equation}
  \footnotesize
  
\InputIfFileExists{middle.tikz}{}{\input{.//tikz//middle.tikz}}

  \normalsize
  \label{equ:middle}
\end{equation}

By application of the spider equality (Section \ref{sec:calculus}) we get the normal form below, which computes a meaning for the sentence as the element-wise multiplication of the vectors composed for the two themes with the vector of the rheme:

\begin{equation}
  \footnotesize
  
\InputIfFileExists{middle-normal.tikz}{}{\input{.//tikz//middle-normal.tikz}}

  \normalsize
  \label{equ:middle-normal}
\end{equation}

\section{Conclusion and future work}

The present paper provides a first account of intonation and information structure for the emerging field of categorical compositional distributional semantics. In a more generic level, it lays the groundwork for a model capable of accommodating two different types of composition over a distributional setting. An experimental evaluation is deferred for the future, preferably in the context of a question-answering task. There is also a lot of interesting work to be done on the theory side. At the current stage, for example, the semantic value of intonational boundaries is given by direct assignment of a specific morphism---a common practice in the past for the relevant literature. A future direction, then, more aligned with the categorical nature of the model, would be to embed the appropriate translation into the functorial passage itself. This challenging goal requires novel theoretical contributions that will elevate the concept of a pregroup grammar to a new entity equipped with Frobenius structure.

Finally, the categorical compositional model of \newcite{piedeleu2015open-calco} is very relevant to our interests, since it can accommodate a variety of non-commutative Frobenius algebras the linguistic intuition of which in relation to this work remains to be explored.

\section*{Acknowledgements}

We would like to thank the three anonymous reviewers for their comments, and Bob Coecke for great discussions on the paper. Financial support from AFOSR is gratefully acknowledged by the authors.

\newpage
\bibliographystyle{acl2012}
\bibliography{refs}

\end{document}

%% file: macros.tex
\pgfdeclarelayer{edgelayer}
\pgfdeclarelayer{nodelayer}
\pgfsetlayers{edgelayer,nodelayer,main}

\tikzstyle{none}=[inner sep=0pt]
\tikzstyle{plain}=[inner sep=0pt]
\tikzstyle{every picture}=[baseline=(current bounding box).east,scale=0.3,node distance=5mm]

\newcommand{\ctikzfig}[1]{%
\begin{center}\rm
  
\InputIfFileExists{#1.tikz}{}{\input{.//tikz//#1.tikz}}

\end{center}}

\newcommand{\ov}{\overrightarrow}
\newcommand{\ol}{\overline}

\newcommand{\ten}{\otimes}

\newcommand{\sem}[1]{\llbracket #1 \rrbracket}

\newcommand{\F}{\mathcal{F}}

\newcommand{\theme}[1]{[#1]$_{\mathsf{T}}$}
\newcommand{\rheme}[1]{[#1]$_{\mathsf{R}}$}

\newtheorem{definition}{Definition}[section]

\newtheorem{postulate}{Postulate}[section]

\makeatletter
\def\maketag@@@#1{\hbox{\m@th\normalfont\normalsize#1}}
\makeatother

%% file: colours.tex
\usepackage{color}

\def\bR{\begin{color}{red}} 
\def\bB{\begin{color}{blue}}
\def\bM{\begin{color}{magenta}}
\def\bC{\begin{color}{cyan}}
\def\bW{\begin{color}{white}}
\def\bBl{\begin{color}{black}} 
\def\bG{\begin{color}{green}}
\def\bY{\begin{color}{yellow}}
\def\e{\end{color}}

%% file: tikz/compact-diag.tikz
\begin{tikzpicture}[scale=0.75]
	\begin{pgfonlayer}{nodelayer}
		\node [style=none] (0) at (-8, 3) {$A$};
		\node [style=none] (1) at (-12.5, 2) {};
		\node [style=none] (2) at (-8, 2) {};
		\node [style=none] (3) at (-9, 1) {};
		\node [style=none] (4) at (-8, 1) {};
		\node [style=none] (5) at (-7, 1) {};
		\node [style=none] (6) at (-11.75, -0) {$A$};
		\node [style=none] (7) at (-8, 0) {$f$};
		\node [style=none] (8) at (-9, -0.75) {};
		\node [style=none] (9) at (-8, -0.75) {};
		\node [style=none] (10) at (-7, -0.75) {};
		\node [style=none] (11) at (-12.5, -2) {};
		\node [style=none] (12) at (-8, -2) {};
		\node [style=none] (13) at (-8, -2.75) {$B$};
	\end{pgfonlayer}
	\begin{pgfonlayer}{edgelayer}
		\draw [thick] (9.center) to (12.center);
		\draw [thick] (1.center) to (11.center);
		\draw [thick] (3.center) to (5.center);
		\draw [thick] (5.center) to (10.center);
		\draw [thick] (10.center) to (8.center);
		\draw [thick] (2.center) to (4.center);
		\draw [thick] (3.center) to (8.center);
	\end{pgfonlayer}
\end{tikzpicture}

%% file: tikz/compact-diag-tensor.tikz
\begin{tikzpicture}[scale=0.75]
	\begin{pgfonlayer}{nodelayer}
		\node [style=none] (0) at (5, 6.5) {$A$};
		\node [style=none] (1) at (5, 5.75) {};
		\node [style=none] (2) at (4, 4.75) {};
		\node [style=none] (3) at (5, 4.75) {};
		\node [style=none] (4) at (6, 4.75) {};
		\node [style=none] (5) at (-2.75, 3.75) {$A$};
		\node [style=none] (6) at (-0.25, 3.75) {$C$};
		\node [style=none] (7) at (5, 3.75) {$f$};
		\node [style=none] (8) at (-10, 3) {};
		\node [style=none] (9) at (-8, 3) {};
		\node [style=none] (10) at (-3, 3) {};
		\node [style=none] (11) at (-0.5, 3) {};
		\node [style=none] (12) at (4, 2.75) {};
		\node [style=none] (13) at (5, 2.75) {};
		\node [style=none] (14) at (6, 2.75) {};
		\node [style=none] (15) at (-4, 2) {};
		\node [style=none] (16) at (-3, 2) {};
		\node [style=none] (17) at (-2, 2) {};
		\node [style=none] (18) at (-1.5, 2) {};
		\node [style=none] (19) at (-0.5, 2) {};
		\node [style=none] (20) at (0.5, 2) {};
		\node [style=none] (21) at (5, 1.75) {};
		\node [style=none] (22) at (-10.75, 1) {$A$};
		\node [style=none] (23) at (-7.25, 1) {$B$};
		\node [style=none] (24) at (-3, 1) {$f$};
		\node [style=none] (25) at (-0.5, 1) {$g$};
		\node [style=none] (26) at (5, 1) {$B$};
		\node [style=none] (27) at (5, 0.25) {};
		\node [style=none] (28) at (-4, 0) {};
		\node [style=none] (29) at (-3, 0) {};
		\node [style=none] (30) at (-2, 0) {};
		\node [style=none] (31) at (-1.5, 0) {};
		\node [style=none] (32) at (-0.5, 0) {};
		\node [style=none] (33) at (0.5, 0) {};
		\node [style=none] (34) at (4, -0.75) {};
		\node [style=none] (35) at (5, -0.75) {};
		\node [style=none] (36) at (6, -0.75) {};
		\node [style=none] (37) at (-10, -1) {};
		\node [style=none] (38) at (-8, -1) {};
		\node [style=none] (39) at (-3, -1) {};
		\node [style=none] (40) at (-0.5, -1) {};
		\node [style=none] (41) at (-3, -1.75) {$B$};
		\node [style=none] (42) at (-0.5, -1.75) {$D$};
		\node [style=none] (43) at (5, -1.75) {$h$};
		\node [style=none] (44) at (4, -2.75) {};
		\node [style=none] (45) at (5, -2.75) {};
		\node [style=none] (46) at (6, -2.75) {};
		\node [style=none] (47) at (5, -3.75) {};
		\node [style=none] (48) at (5, -4.5) {$C$};
	\end{pgfonlayer}
	\begin{pgfonlayer}{edgelayer}
		\draw [thick] (1.center) to (3.center);
		\draw [thick] (30.center) to (28.center);
		\draw [thick] (34.center) to (36.center);
		\draw [thick] (18.center) to (20.center);
		\draw [thick] (45.center) to (47.center);
		\draw [thick] (14.center) to (12.center);
		\draw [thick] (17.center) to (30.center);
		\draw [thick] (18.center) to (31.center);
		\draw [thick] (27.center) to (35.center);
		\draw [thick] (11.center) to (19.center);
		\draw [thick] (4.center) to (14.center);
		\draw [thick] (15.center) to (17.center);
		\draw [thick] (34.center) to (44.center);
		\draw [thick] (2.center) to (12.center);
		\draw [thick] (29.center) to (39.center);
		\draw [thick] (46.center) to (44.center);
		\draw [thick] (33.center) to (31.center);
		\draw [thick] (9.center) to (38.center);
		\draw [thick] (2.center) to (4.center);
		\draw [thick] (10.center) to (16.center);
		\draw [thick] (32.center) to (40.center);
		\draw [thick] (13.center) to (21.center);
		\draw [thick] (36.center) to (46.center);
		\draw [thick] (20.center) to (33.center);
		\draw [thick] (8.center) to (37.center);
		\draw [thick] (15.center) to (28.center);
	\end{pgfonlayer}
\end{tikzpicture}

%% file: tikz/compact-cap-cup.tikz
\begin{tikzpicture}[scale=0.75]
	\begin{pgfonlayer}{nodelayer}
		\node [style=none] (0) at (-5, 0) {};
		\node [style=none] (1) at (-2, 0) {};
		\node [style=none] (2) at (-5, 0.75) {$A^l$};
		\node [style=none] (3) at (2, -0.75) {$A$};
		\node [style=none] (4) at (5, -0.75) {$A^l$};
		\node [style=none] (5) at (2, 0) {};
		\node [style=none] (6) at (5, 0) {};
		\node [style=none] (7) at (-2, 0.75) {$A$};
	\end{pgfonlayer}
	\begin{pgfonlayer}{edgelayer}
		\draw [thick, bend right=90, looseness=1.50] (0.center) to (1.center);
		\draw [thick, bend left=90, looseness=1.75] (5.center) to (6.center);
	\end{pgfonlayer}
\end{tikzpicture}

%% file: tikz/compact-yank.tikz
\begin{tikzpicture}[scale=0.75]
	\begin{pgfonlayer}{nodelayer}
		\node [style=none] (0) at (-5, 2.75) {};
		\node [style=none] (1) at (5, 2.5) {};
		\node [style=none] (2) at (-5, 1.25) {};
		\node [style=none] (3) at (-2, 1.25) {};
		\node [style=none] (4) at (1, 1.25) {};
		\node [style=none] (5) at (-5, 0.5) {$A^l$};
		\node [style=none] (6) at (-2, 0.5) {$A$};
		\node [style=none] (7) at (1, 0.5) {$A^l$};
		\node [style=none] (8) at (5.75, 0.5) {$A$};
		\node [style=none] (9) at (3, 0) {$=$};
		\node [style=none] (10) at (-5, -0.25) {};
		\node [style=none] (11) at (-2, -0.25) {};
		\node [style=none] (12) at (1, -0.25) {};
		\node [style=none] (13) at (5, -1.5) {};
		\node [style=none] (14) at (1, -1.75) {};
	\end{pgfonlayer}
	\begin{pgfonlayer}{edgelayer}
		\draw [thick, bend right=90, looseness=1.50] (10.center) to (11.center);
		\draw [thick] (12.center) to (14.center);
		\draw [thick] (1.center) to (13.center);
		\draw [thick, bend left=90, looseness=1.75] (3.center) to (4.center);
		\draw [thick] (0.center) to (2.center);
	\end{pgfonlayer}
\end{tikzpicture}

%% file: tikz/spider.tikz
\begin{tikzpicture}[baseline=0 pt]
	\begin{pgfonlayer}{nodelayer}
		\node [style=none] (0) at (-4.75, 4.5) {};
		\node [style=none] (1) at (-3.25, 4.5) {};
		\node [style=none] (2) at (-0.5, 4.5) {};
		\node [draw, style=none, minimum size=0.12 cm, circle, fill=white] (3) at (-4, 3.6) {};
		\node [style=none] (4) at (-2.5, 3.6) {};
		\node [draw, style=none, minimum size=0.12 cm, circle, fill=white] (5) at (-3.25, 2.7) {};
		\node [style=none] (6) at (-2.5, 2) {$\ddots$};
		\node [style=none] (7) at (1.25, 1.75) {};
		\node [style=none] (8) at (2, 1.75) {};
		\node [style=none] (9) at (3, 1.75) {$\cdots$};
		\node [style=none] (10) at (3.75, 1.75) {};
		\node [style=none] (11) at (-2, 1) {};
		\node [style=none] (12) at (0, 0.5) {$=$};
		\node [draw, style=none, minimum size=0.12 cm, circle, fill=white] (13) at (2.5, 0.4) {};
		\node [style=none] (14) at (-2, -0) {};
		\node [style=none] (15) at (-2.5, -0.75) {$\ddots$};
		\node [style=none] (16) at (1.25, -0.75) {};
		\node [style=none] (17) at (2, -0.75) {};
		\node [style=none] (18) at (3, -0.75) {$\cdots$};
		\node [style=none] (19) at (3.75, -0.75) {};
		\node [draw, style=none, minimum size=0.12 cm, circle, fill=white] (20) at (-3.25, -1.8) {};
		\node [draw, style=none, minimum size=0.12 cm, circle, fill=white] (21) at (-4, -2.7) {};
		\node [style=none] (22) at (-2.5, -2.7) {};
		\node [style=none] (23) at (-4.75, -3.5) {};
		\node [style=none] (24) at (-3.25, -3.5) {};
		\node [style=none] (25) at (-0.5, -3.5) {};
	\end{pgfonlayer}
	\begin{pgfonlayer}{edgelayer}
		\draw [thick, bend left=90, looseness=1.50] (16.center) to (19.center);
		\draw [thick, bend left=90, looseness=2.00] (23.center) to (24.center);
		\draw [thick, bend right, looseness=0.75] (11.center) to (2.center);
		\draw [thick, bend left, looseness=0.75] (14.center) to (25.center);
		\draw [thick, bend left=270, looseness=1.75] (7.center) to (10.center);
		\draw [thick, bend right=90, looseness=2.00] (3.center) to (4.center);
		\draw [thick, bend right] (8.center) to (13.center);
		\draw [thick, bend left] (17.center) to (13.center);
		\draw [thick, bend right=90, looseness=2.00] (0.center) to (1.center);
		\draw [thick, bend left=90, looseness=2.00] (21.center) to (22.center);
	\end{pgfonlayer}
\end{tikzpicture}

%% file: tikz/dtrans-preg.tikz
\begin{tikzpicture}
	\begin{pgfonlayer}{nodelayer}
		\node [style=none, text depth=0.25 ex, text height=1.5 ex] (0) at (-5.75, 2.75) {John};
		\node [style=none, text depth=0.25 ex, text height=1.5 ex] (1) at (-2, 2.75) {gave};
		\node [style=none, text depth=0.25 ex, text height=1.5 ex] (2) at (1.75, 2.75) {Mary};
		\node [style=none, text depth=0.25 ex, text height=1.5 ex] (3) at (5.5, 2.75) {a flower};
		\node [style=none, text depth=0.25 ex, text height=1.5 ex] (4) at (-5.75, 1.25) {$n$};
		\node [style=none, text depth=0.25 ex, text height=1.5 ex] (5) at (-3.25, 1.25) {$n^r$};
		\node [style=none, text depth=0.25 ex, text height=1.5 ex] (6) at (-2.5, 1.25) {$s$};
		\node [style=none, text depth=0.25 ex, text height=1.5 ex] (7) at (-1.5, 1.25) {$n^l$};
		\node [style=none, text depth=0.25 ex, text height=1.5 ex] (8) at (-0.75, 1.25) {$n^l$};
		\node [style=none, text depth=0.25 ex, text height=1.5 ex] (9) at (1.75, 1.25) {$n$};
		\node [style=none, text depth=0.25 ex, text height=1.5 ex] (10) at (5.5, 1.25) {$n$};
		\node [style=none] (11) at (-5.75, 0.5) {};
		\node [style=none] (12) at (-3.25, 0.5) {};
		\node [style=none] (13) at (-2.5, 0.5) {};
		\node [style=none] (14) at (-1.5, 0.5) {};
		\node [style=none] (15) at (-0.75, 0.5) {};
		\node [style=none] (16) at (1.75, 0.5) {};
		\node [style=none] (17) at (5.5, 0.5) {};
		\node [style=none] (18) at (-2.5, -1.25) {};
	\end{pgfonlayer}
	\begin{pgfonlayer}{edgelayer}
		\draw [thick, bend right=90, looseness=1.25] (11.center) to (12.center);
		\draw [thick, bend right=90, looseness=1.25] (15.center) to (16.center);
		\draw [thick] (13.center) to (18.center);
		\draw [thick, bend right=90, looseness=0.75] (14.center) to (17.center);
	\end{pgfonlayer}
\end{tikzpicture}

%% file: tikz/inton-preg.tikz
\begin{tikzpicture}
	\begin{pgfonlayer}{nodelayer}
		\node [text height=1.5 ex, text depth=0.25 ex, style=none] (0) at (-7.25, 3) {Mary};
		\node [text height=1.5 ex, text depth=0.25 ex, style=none] (1) at (-4, 3) {likes};
		\node [style=none] (2) at (0.25, 3) {$\triangleright$};
		\node [text height=1.5 ex, text depth=0.25 ex, style=none] (3) at (4.5, 3) {musicals};
		\node [text height=1.5 ex, text depth=0.25 ex, style=none] (4) at (-7.25, 1.25) {$n$};
		\node [text height=1.5 ex, text depth=0.25 ex, style=none] (5) at (-4.75, 1.25) {$n^r$};
		\node [text height=1.5 ex, text depth=0.25 ex, style=none] (6) at (-3.25, 1.25) {$\theta$};
		\node [text height=1.5 ex, text depth=0.25 ex, style=none] (7) at (-0.75, 1.25) {$\theta^r$};
		\node [text height=1.5 ex, text depth=0.25 ex, style=none, style=none] (8) at (0.25, 1.25) {$s$};
		\node [text height=1.5 ex, text depth=0.25 ex, style=none] (9) at (1.25, 1.25) {$\rho^l$};
		\node [text height=1.5 ex, text depth=0.25 ex, style=none] (10) at (4.5, 1.25) {$\rho$};
		\node [style=none] (11) at (-7.25, 0.5) {};
		\node [style=none] (12) at (-4.75, 0.5) {};
		\node [style=none] (13) at (-3.25, 0.5) {};
		\node [style=none] (14) at (-0.75, 0.5) {};
		\node [style=none] (15) at (0.25, 0.5) {};
		\node [style=none] (16) at (1.25, 0.5) {};
		\node [style=none] (17) at (4.5, 0.5) {};
		\node [style=none] (18) at (0.25, -1.25) {};
	\end{pgfonlayer}
	\begin{pgfonlayer}{edgelayer}
		\draw [thick, bend left=270, looseness=1.25] (11.center) to (12.center);
		\draw [thick] (15.center) to (18.center);
		\draw [thick, bend left=270] (16.center) to (17.center);
		\draw [thick, bend left=270, looseness=1.25] (13.center) to (14.center);
	\end{pgfonlayer}
\end{tikzpicture}

%% file: tikz/inton-preg2.tikz
\begin{tikzpicture}
	\begin{pgfonlayer}{nodelayer}
		\node [text height=1.5 ex, text depth=0.25 ex, style=none] (0) at (-7.25, 3) {Mary};
		\node [style=none] (1) at (-4, 3) {$\triangleleft$};
		\node [text height=1.5 ex, text depth=0.25 ex, style=none] (2) at (0.25, 3) {likes};
		\node [text height=1.5 ex, text depth=0.25 ex, style=none] (3) at (4, 3) {musicals};
		\node [text height=1.5 ex, text depth=0.25 ex, style=none] (4) at (-7.25, 1.5) {$\rho$};
		\node [text height=1.5 ex, text depth=0.25 ex, style=none] (5) at (-4.75, 1.5) {$\rho^r$};
		\node [text height=1.5 ex, text depth=0.25 ex, style=none, style=none] (6) at (-3.75, 1.5) {$s$};
		\node [text height=1.5 ex, text depth=0.25 ex, style=none] (7) at (-2.75, 1.5) {$\theta^l$};
		\node [text height=1.5 ex, text depth=0.25 ex, style=none] (8) at (-0.25, 1.5) {$\theta$};
		\node [text height=1.5 ex, text depth=0.25 ex, style=none] (9) at (0.75, 1.5) {$n^l$};
		\node [text height=1.5 ex, text depth=0.25 ex, style=none] (10) at (4, 1.5) {$n$};
		\node [style=none] (11) at (-7.25, 0.5) {};
		\node [style=none] (12) at (-4.75, 0.5) {};
		\node [style=none] (13) at (-3.75, 0.5) {};
		\node [style=none] (14) at (-2.75, 0.5) {};
		\node [style=none] (15) at (-0.25, 0.5) {};
		\node [style=none] (16) at (0.75, 0.5) {};
		\node [style=none] (17) at (4, 0.5) {};
		\node [style=none] (18) at (-3.75, -1.25) {};
	\end{pgfonlayer}
	\begin{pgfonlayer}{edgelayer}
		\draw [thick, bend left=270, looseness=1.25] (11.center) to (12.center);
		\draw [thick] (13.center) to (18.center);
		\draw [thick, bend left=270] (16.center) to (17.center);
		\draw [thick, bend left=270, looseness=1.25] (14.center) to (15.center);
	\end{pgfonlayer}
\end{tikzpicture}

%% file: tikz/inton-fvect.tikz
\begin{tikzpicture}
	\begin{pgfonlayer}{nodelayer}
		\node [text height=1.5 ex, text depth=0.25 ex, style=none] (0) at (-7.25, 5.5) {Mary};
		\node [text height=1.5 ex, text depth=0.25 ex, style=none] (1) at (-4, 5.5) {likes};
		\node [style=none] (2) at (0.25, 5.5) {$\triangleright$};
		\node [text height=1.5 ex, text depth=0.25 ex, style=none] (3) at (4.5, 5.5) {musicals};
		\node [style=none] (4) at (0.25, 4.5) {};
		\node [style=none] (5) at (-4, 4.25) {};
		\node [style=none] (6) at (-7.25, 3.75) {};
		\node [style=none] (7) at (4.5, 3.75) {};
		\node [style=none] (8) at (-8, 2.75) {};
		\node [style=none] (9) at (-7.25, 2.75) {};
		\node [style=none] (10) at (-6.5, 2.75) {};
		\node [style=none] (11) at (-5.25, 2.75) {};
		\node [style=none] (12) at (-4.75, 2.75) {};
		\node [style=none] (13) at (-3.25, 2.75) {};
		\node [style=none] (14) at (-2.75, 2.75) {};
		\node [style=none] (15) at (-1.5, 2.75) {};
		\node [style=none] (16) at (-0.75, 2.75) {};
		\node [style=none] (17) at (0.25, 2.75) {};
		\node [style=none] (18) at (1.25, 2.75) {};
		\node [style=none] (19) at (2, 2.75) {};
		\node [style=none] (20) at (3.75, 2.75) {};
		\node [style=none] (21) at (4.5, 2.75) {};
		\node [style=none] (22) at (5.25, 2.75) {};
		\node [style=none] (23) at (-7.25, 2) {};
		\node [style=none] (24) at (-4.75, 2) {};
		\node [style=none] (25) at (-3.25, 2) {};
		\node [style=none] (26) at (-0.75, 2) {};
		\node [style=none] (27) at (0.25, 2) {};
		\node [style=none] (28) at (1.25, 2) {};
		\node [style=none] (29) at (4.5, 2) {};
		\node [text height=1.5 ex, text depth=0.25 ex, style=none] (30) at (-7.25, 1.25) {$N$};
		\node [text height=1.5 ex, text depth=0.25 ex, style=none] (31) at (-4.75, 1.25) {$N^r$};
		\node [text height=1.5 ex, text depth=0.25 ex, style=none] (32) at (-3.25, 1.25) {$\Theta$};
		\node [text height=1.5 ex, text depth=0.25 ex, style=none] (33) at (-0.75, 1.25) {$\Theta^r$};
		\node [text height=1.5 ex, text depth=0.25 ex, style=none, style=none] (34) at (0.25, 1.25) {$S$};
		\node [text height=1.5 ex, text depth=0.25 ex, style=none] (35) at (1.25, 1.25) {$P^l$};
		\node [text height=1.5 ex, text depth=0.25 ex, style=none] (36) at (4.5, 1.25) {$P$};
		\node [style=none] (37) at (-7.25, 0.5) {};
		\node [style=none] (38) at (-4.75, 0.5) {};
		\node [style=none] (39) at (-3.25, 0.5) {};
		\node [style=none] (40) at (-0.75, 0.5) {};
		\node [style=none] (41) at (0.25, 0.5) {};
		\node [style=none] (42) at (1.25, 0.5) {};
		\node [style=none] (43) at (4.5, 0.5) {};
		\node [style=none] (44) at (0.25, -1.25) {};
	\end{pgfonlayer}
	\begin{pgfonlayer}{edgelayer}
		\draw [thick] (16.center) to (26.center);
		\draw [thick] (20.center) to (22.center);
		\draw [thick] (21.center) to (29.center);
		\draw [thick, bend left=270] (42.center) to (43.center);
		\draw [thick] (15.center) to (4.center);
		\draw [thick] (11.center) to (5.center);
		\draw [thick, bend left=270, looseness=1.25] (39.center) to (40.center);
		\draw [thick, bend left=270, looseness=1.25] (37.center) to (38.center);
		\draw [thick] (6.center) to (10.center);
		\draw [thick] (20.center) to (7.center);
		\draw [thick] (7.center) to (22.center);
		\draw [thick] (15.center) to (19.center);
		\draw [thick] (11.center) to (14.center);
		\draw [thick] (9.center) to (23.center);
		\draw [thick] (41.center) to (44.center);
		\draw [thick] (12.center) to (24.center);
		\draw [thick] (18.center) to (28.center);
		\draw [thick] (17.center) to (27.center);
		\draw [thick] (8.center) to (6.center);
		\draw [thick] (8.center) to (10.center);
		\draw [thick] (13.center) to (25.center);
		\draw [thick] (5.center) to (14.center);
		\draw [thick] (4.center) to (19.center);
	\end{pgfonlayer}
\end{tikzpicture}

%% file: tikz/boundary1.tikz
\begin{tikzpicture}
	\begin{pgfonlayer}{nodelayer}
		\node [style=none, text height=1.5 ex, text depth=0.25 ex] (0) at (21.75, 3.75) {$\triangleright$};
		\node [style=none] (1) at (21.75, 3) {};
		\node [style=none, text height=1.5 ex, text depth=0.25 ex] (2) at (31, 3) {theme};
		\node [style=none, text height=1.5 ex, text depth=0.25 ex] (3) at (34, 3) {rheme};
		\node [style=none, text height=1.5 ex, text depth=0.25 ex] (4) at (16.75, 2.5) {theme};
		\node [style=none, text height=1.5 ex, text depth=0.25 ex] (5) at (26.75, 2.5) {rheme};
		\node [style=none] (6) at (31, 2) {};
		\node [style=none] (7) at (34, 2) {};
		\node [style=none] (8) at (16.75, 1.5) {};
		\node [style=none] (9) at (26.75, 1.5) {};
		\node [style=none] (10) at (20.5, 1.25) {};
		\node [style=none] (11) at (21.25, 1.25) {};
		\node [style=none] (12) at (22.25, 1.25) {};
		\node [style=none] (13) at (23, 1.25) {};
		\node [style=none] (14) at (30, 1) {};
		\node [style=none] (15) at (31, 1) {};
		\node [style=none] (16) at (32, 1) {};
		\node [style=none] (17) at (33, 1) {};
		\node [style=none] (18) at (34, 1) {};
		\node [style=none] (19) at (35, 1) {};
		\node [draw, circle, minimum size=0.12 cm, fill=white, style=none] (20) at (21.75, 0.85) {};
		\node [style=none] (21) at (15.75, 0.5) {};
		\node [style=none] (22) at (16.75, 0.5) {};
		\node [style=none] (23) at (17.75, 0.5) {};
		\node [style=none] (24) at (19.25, 0.5) {};
		\node [style=none] (25) at (20.5, 0.5) {};
		\node [style=none] (26) at (23, 0.5) {};
		\node [style=none] (27) at (24.25, 0.5) {};
		\node [style=none] (28) at (25.75, 0.5) {};
		\node [style=none] (29) at (26.75, 0.5) {};
		\node [style=none] (30) at (27.75, 0.5) {};
		\node [style=none] (31) at (31, 0.25) {};
		\node [style=none] (32) at (34, 0.25) {};
		\node [style=none] (33) at (16.75, -0.25) {};
		\node [style=none] (34) at (20.5, -0.25) {};
		\node [style=none] (35) at (21.75, -0.25) {};
		\node [style=none] (36) at (23, -0.25) {};
		\node [style=none] (37) at (26.75, -0.25) {};
		\node [style=none, text height=1.5 ex, text depth=0.25 ex] (38) at (31, -0.25) {\footnotesize{$W$}};
		\node [style=none, text height=1.5 ex, text depth=0.25 ex] (39) at (34, -0.25) {\footnotesize{$W$}};
		\node [style=none, text height=1.5 ex, text depth=0.25 ex] (40) at (29, -0.5) {$=$};
		\node [style=none, text height=1.5 ex, text depth=0.25 ex] (41) at (16.75, -0.75) {\footnotesize{$W$}};
		\node [style=none, text height=1.5 ex, text depth=0.25 ex] (42) at (20.5, -0.75) {\footnotesize{$W$}};
		\node [style=none, text height=1.5 ex, text depth=0.25 ex] (43) at (21.75, -0.75) {\footnotesize{$W$}};
		\node [style=none, text height=1.5 ex, text depth=0.25 ex] (44) at (23, -0.75) {\footnotesize{$W$}};
		\node [style=none, text height=1.5 ex, text depth=0.25 ex] (45) at (26.75, -0.75) {\footnotesize{$W$}};
		\node [style=none] (46) at (31, -0.75) {};
		\node [style=none] (47) at (34, -0.75) {};
		\node [style=none] (48) at (16.75, -1.25) {};
		\node [style=none] (49) at (20.5, -1.25) {};
		\node [style=none] (50) at (21.75, -1.25) {};
		\node [style=none] (51) at (23, -1.25) {};
		\node [style=none] (52) at (26.75, -1.25) {};
		\node [draw, circle, minimum size=0.15 cm, fill=white, style=none] (53) at (32.5, -1.65) {};
		\node [style=none] (54) at (21.75, -2.5) {};
		\node [style=none] (55) at (32.5, -2.75) {};
	\end{pgfonlayer}
	\begin{pgfonlayer}{edgelayer}
		\draw [thick, looseness=0.00] (7.center) to (19.center);
		\draw [thick] (10.center) to (25.center);
		\draw [thick, looseness=0.00] (28.center) to (9.center);
		\draw [thick, bend left=90, looseness=2.25] (12.center) to (13.center);
		\draw [thick, looseness=0.00] (17.center) to (19.center);
		\draw [thick, looseness=0.00] (6.center) to (16.center);
		\draw [thick, looseness=0.00] (14.center) to (6.center);
		\draw [thick, looseness=0.00] (18.center) to (32.center);
		\draw [thick, looseness=0.00] (24.center) to (1.center);
		\draw [thick, looseness=0.00] (29.center) to (37.center);
		\draw [thick, looseness=0.00] (28.center) to (30.center);
		\draw [thick, looseness=0.00] (15.center) to (31.center);
		\draw [thick, bend right=90] (46.center) to (47.center);
		\draw [thick, looseness=0.00] (8.center) to (23.center);
		\draw [thick, bend left=270, looseness=1.50] (11.center) to (12.center);
		\draw [thick, looseness=0.00] (1.center) to (27.center);
		\draw [thick, looseness=0.00] (17.center) to (7.center);
		\draw [thick, looseness=0.00] (9.center) to (30.center);
		\draw [thick, looseness=0.00] (14.center) to (16.center);
		\draw [thick] (53.center) to (55.center);
		\draw [thick, looseness=0.00] (26.center) to (36.center);
		\draw [thick, looseness=0.00] (21.center) to (8.center);
		\draw [thick, looseness=0.00] (22.center) to (33.center);
		\draw [thick, bend left=90, looseness=2.25] (10.center) to (11.center);
		\draw [thick] (13.center) to (26.center);
		\draw [thick] (50.center) to (54.center);
		\draw [thick, looseness=0.00] (24.center) to (27.center);
		\draw [thick, looseness=0.00] (20.center) to (35.center);
		\draw [thick, bend left=270] (51.center) to (52.center);
		\draw [thick, looseness=0.00] (25.center) to (34.center);
		\draw [thick, looseness=0.00] (21.center) to (23.center);
		\draw [thick, bend left=270] (48.center) to (49.center);
	\end{pgfonlayer}
\end{tikzpicture}

%% file: tikz/boundary2.tikz
\begin{tikzpicture}
	\begin{pgfonlayer}{nodelayer}
		\node [style=none, text height=1.5 ex, text depth=0.25 ex] (0) at (21.75, 3.75) {$\triangleleft$};
		\node [style=none] (1) at (21.75, 3) {};
		\node [style=none, text height=1.5 ex, text depth=0.25 ex] (2) at (31, 3) {rheme};
		\node [style=none, text height=1.5 ex, text depth=0.25 ex] (3) at (34, 3) {theme};
		\node [style=none, text height=1.5 ex, text depth=0.25 ex] (4) at (16.75, 2.5) {rheme};
		\node [style=none, text height=1.5 ex, text depth=0.25 ex] (5) at (26.75, 2.5) {theme};
		\node [style=none] (6) at (31, 2) {};
		\node [style=none] (7) at (34, 2) {};
		\node [style=none] (8) at (16.75, 1.5) {};
		\node [style=none] (9) at (26.75, 1.5) {};
		\node [style=none] (10) at (20.5, 1.25) {};
		\node [style=none] (11) at (21.25, 1.25) {};
		\node [style=none] (12) at (22.25, 1.25) {};
		\node [style=none] (13) at (23, 1.25) {};
		\node [style=none] (14) at (30, 1) {};
		\node [style=none] (15) at (31, 1) {};
		\node [style=none] (16) at (32, 1) {};
		\node [style=none] (17) at (33, 1) {};
		\node [style=none] (18) at (34, 1) {};
		\node [style=none] (19) at (35, 1) {};
		\node [draw, circle, minimum size=0.12 cm, fill=white, style=none] (20) at (21.75, 0.85) {};
		\node [style=none] (21) at (15.75, 0.5) {};
		\node [style=none] (22) at (16.75, 0.5) {};
		\node [style=none] (23) at (17.75, 0.5) {};
		\node [style=none] (24) at (19.25, 0.5) {};
		\node [style=none] (25) at (20.5, 0.5) {};
		\node [style=none] (26) at (23, 0.5) {};
		\node [style=none] (27) at (24.25, 0.5) {};
		\node [style=none] (28) at (25.75, 0.5) {};
		\node [style=none] (29) at (26.75, 0.5) {};
		\node [style=none] (30) at (27.75, 0.5) {};
		\node [style=none] (31) at (31, 0.25) {};
		\node [style=none] (32) at (34, 0.25) {};
		\node [style=none] (33) at (16.75, -0.25) {};
		\node [style=none] (34) at (20.5, -0.25) {};
		\node [style=none] (35) at (21.75, -0.25) {};
		\node [style=none] (36) at (23, -0.25) {};
		\node [style=none] (37) at (26.75, -0.25) {};
		\node [style=none, text height=1.5 ex, text depth=0.25 ex] (38) at (31, -0.25) {\footnotesize{$W$}};
		\node [style=none, text height=1.5 ex, text depth=0.25 ex] (39) at (34, -0.25) {\footnotesize{$W$}};
		\node [style=none, text height=1.5 ex, text depth=0.25 ex] (40) at (29, -0.5) {$=$};
		\node [style=none, text height=1.5 ex, text depth=0.25 ex] (41) at (16.75, -0.75) {\footnotesize{$W$}};
		\node [style=none, text height=1.5 ex, text depth=0.25 ex] (42) at (20.5, -0.75) {\footnotesize{$W$}};
		\node [style=none, text height=1.5 ex, text depth=0.25 ex] (43) at (21.75, -0.75) {\footnotesize{$W$}};
		\node [style=none, text height=1.5 ex, text depth=0.25 ex] (44) at (23, -0.75) {\footnotesize{$W$}};
		\node [style=none, text height=1.5 ex, text depth=0.25 ex] (45) at (26.75, -0.75) {\footnotesize{$W$}};
		\node [style=none] (46) at (31, -0.75) {};
		\node [style=none] (47) at (34, -0.75) {};
		\node [style=none] (48) at (16.75, -1.25) {};
		\node [style=none] (49) at (20.5, -1.25) {};
		\node [style=none] (50) at (21.75, -1.25) {};
		\node [style=none] (51) at (23, -1.25) {};
		\node [style=none] (52) at (26.75, -1.25) {};
		\node [draw, circle, minimum size=0.15 cm, fill=white, style=none] (53) at (32.5, -1.65) {};
		\node [style=none] (54) at (21.75, -2.5) {};
		\node [style=none] (55) at (32.5, -2.75) {};
	\end{pgfonlayer}
	\begin{pgfonlayer}{edgelayer}
		\draw [thick, looseness=0.00] (7.center) to (19.center);
		\draw [thick] (10.center) to (25.center);
		\draw [thick, looseness=0.00] (28.center) to (9.center);
		\draw [thick, bend left=90, looseness=2.25] (12.center) to (13.center);
		\draw [thick, looseness=0.00] (17.center) to (19.center);
		\draw [thick, looseness=0.00] (6.center) to (16.center);
		\draw [thick, looseness=0.00] (14.center) to (6.center);
		\draw [thick, looseness=0.00] (18.center) to (32.center);
		\draw [thick, looseness=0.00] (24.center) to (1.center);
		\draw [thick, looseness=0.00] (29.center) to (37.center);
		\draw [thick, looseness=0.00] (28.center) to (30.center);
		\draw [thick, looseness=0.00] (15.center) to (31.center);
		\draw [thick, bend right=90] (46.center) to (47.center);
		\draw [thick, looseness=0.00] (8.center) to (23.center);
		\draw [thick, bend left=270, looseness=1.50] (11.center) to (12.center);
		\draw [thick, looseness=0.00] (1.center) to (27.center);
		\draw [thick, looseness=0.00] (17.center) to (7.center);
		\draw [thick, looseness=0.00] (9.center) to (30.center);
		\draw [thick, looseness=0.00] (14.center) to (16.center);
		\draw [thick] (53.center) to (55.center);
		\draw [thick, looseness=0.00] (26.center) to (36.center);
		\draw [thick, looseness=0.00] (21.center) to (8.center);
		\draw [thick, looseness=0.00] (22.center) to (33.center);
		\draw [thick, bend left=90, looseness=2.25] (10.center) to (11.center);
		\draw [thick] (13.center) to (26.center);
		\draw [thick] (50.center) to (54.center);
		\draw [thick, looseness=0.00] (24.center) to (27.center);
		\draw [thick, looseness=0.00] (20.center) to (35.center);
		\draw [thick, bend left=270] (51.center) to (52.center);
		\draw [thick, looseness=0.00] (25.center) to (34.center);
		\draw [thick, looseness=0.00] (21.center) to (23.center);
		\draw [thick, bend left=270] (48.center) to (49.center);
	\end{pgfonlayer}
\end{tikzpicture}

%% file: tikz/frob-mu.tikz
\begin{tikzpicture}
	\begin{pgfonlayer}{nodelayer}
		\node [style=none, text height=1.5 ex, text depth=0.25 ex] (0) at (31, 3) {$\overrightarrow{v_1}$};
		\node [style=none, text height=1.5 ex, text depth=0.25 ex] (1) at (34, 3) {$\overrightarrow{v_2}$};
		\node [style=none] (2) at (31, 2) {};
		\node [style=none] (3) at (34, 2) {};
		\node [style=none] (4) at (30, 1) {};
		\node [style=none] (5) at (31, 1) {};
		\node [style=none] (6) at (32, 1) {};
		\node [style=none] (7) at (33, 1) {};
		\node [style=none] (8) at (34, 1) {};
		\node [style=none] (9) at (35, 1) {};
		\node [style=none] (10) at (31, 0.25) {};
		\node [style=none] (11) at (34, 0.25) {};
		\node [style=none] (12) at (23.5, -0.25) {$\mu( \ov{v_1} \ten \ov{v_2}) =  \ov{v_1} \odot \ov{v_2} = $};
		\node [style=none, text height=1.5 ex, text depth=0.25 ex] (13) at (31, -0.25) {\footnotesize{$V$}};
		\node [style=none, text height=1.5 ex, text depth=0.25 ex] (14) at (34, -0.25) {\footnotesize{$V$}};
		\node [style=none] (15) at (31, -0.75) {};
		\node [style=none] (16) at (34, -0.75) {};
		\node [draw, circle, minimum size=0.15 cm, fill=white, style=none] (17) at (32.5, -1.65) {};
		\node [style=none] (18) at (32.5, -2.75) {};
	\end{pgfonlayer}
	\begin{pgfonlayer}{edgelayer}
		\draw [thick, looseness=0.00] (4.center) to (6.center);
		\draw [thick, looseness=0.00] (3.center) to (9.center);
		\draw [thick, looseness=0.00] (2.center) to (6.center);
		\draw [thick, looseness=0.00] (7.center) to (9.center);
		\draw [thick, bend right=90] (15.center) to (16.center);
		\draw [thick] (17.center) to (18.center);
		\draw [thick, looseness=0.00] (7.center) to (3.center);
		\draw [thick, looseness=0.00] (8.center) to (11.center);
		\draw [thick, looseness=0.00] (4.center) to (2.center);
		\draw [thick, looseness=0.00] (5.center) to (10.center);
	\end{pgfonlayer}
\end{tikzpicture}

%% file: tikz/into-der1.tikz
\begin{tikzpicture}
	\begin{pgfonlayer}{nodelayer}
		\node [style=none] (0) at (0.25, 5.75) {$\triangleright$};
		\node [text height=1.5 ex, text depth=0.25 ex, style=none] (1) at (-7.25, 5.5) {Mary};
		\node [text height=1.5 ex, text depth=0.25 ex, style=none] (2) at (-4, 5.5) {likes};
		\node [text height=1.5 ex, text depth=0.25 ex, style=none] (3) at (4.5, 5.5) {musicals};
		\node [text height=1.5 ex, text depth=0.25 ex, style=none] (4) at (8.5, 5.5) {Mary};
		\node [text height=1.5 ex, text depth=0.25 ex, style=none] (5) at (11.5, 5.5) {likes};
		\node [text height=1.5 ex, text depth=0.25 ex, style=none] (6) at (15, 5.5) {musicals};
		\node [style=none] (7) at (0.25, 5.25) {};
		\node [style=none] (8) at (-4, 4.25) {};
		\node [style=none] (9) at (11.5, 4.25) {};
		\node [style=none] (10) at (-7.25, 3.75) {};
		\node [style=none] (11) at (4.5, 3.75) {};
		\node [style=none] (12) at (8.5, 3.75) {};
		\node [style=none] (13) at (14.75, 3.75) {};
		\node [style=none] (14) at (-1, 3.5) {};
		\node [style=none] (15) at (-0.25, 3.5) {};
		\node [style=none] (16) at (0.75, 3.5) {};
		\node [style=none] (17) at (1.5, 3.5) {};
		\node [draw, circle, minimum size=0.12 cm, fill=white, style=none] (18) at (0.25, 3) {};
		\node [style=none] (19) at (-8, 2.75) {};
		\node [style=none] (20) at (-7.25, 2.75) {};
		\node [style=none] (21) at (-6.5, 2.75) {};
		\node [style=none] (22) at (-5.25, 2.75) {};
		\node [style=none] (23) at (-4.75, 2.75) {};
		\node [style=none] (24) at (-3.25, 2.75) {};
		\node [style=none] (25) at (-2.75, 2.75) {};
		\node [style=none] (26) at (-2.25, 2.75) {};
		\node [style=none] (27) at (-1, 2.75) {};
		\node [style=none] (28) at (-1, 2.75) {};
		\node [style=none] (29) at (0.25, 2.75) {};
		\node [style=none] (30) at (0.25, 2.75) {};
		\node [style=none] (31) at (1.5, 2.75) {};
		\node [style=none] (32) at (1.5, 2.75) {};
		\node [style=none] (33) at (2.75, 2.75) {};
		\node [style=none] (34) at (3.75, 2.75) {};
		\node [style=none] (35) at (4.5, 2.75) {};
		\node [style=none] (36) at (5.25, 2.75) {};
		\node [style=none] (37) at (7.75, 2.75) {};
		\node [style=none] (38) at (8.5, 2.75) {};
		\node [style=none] (39) at (9.25, 2.75) {};
		\node [style=none] (40) at (10.25, 2.75) {};
		\node [style=none] (41) at (10.75, 2.75) {};
		\node [style=none] (42) at (12.25, 2.75) {};
		\node [style=none] (43) at (12.75, 2.75) {};
		\node [style=none] (44) at (14, 2.75) {};
		\node [style=none] (45) at (14.75, 2.75) {};
		\node [style=none] (46) at (15.5, 2.75) {};
		\node [style=none] (47) at (-7.25, 2) {};
		\node [style=none] (48) at (-4.75, 2) {};
		\node [style=none] (49) at (-3.25, 2) {};
		\node [style=none] (50) at (-1, 2) {};
		\node [style=none] (51) at (0.25, 2) {};
		\node [style=none] (52) at (1.5, 2) {};
		\node [style=none] (53) at (4.5, 2) {};
		\node [style=none] (54) at (6.5, 2) {$=$};
		\node [style=none] (55) at (8.5, 2) {};
		\node [style=none] (56) at (10.75, 2) {};
		\node [style=none] (57) at (12.25, 2) {};
		\node [style=none] (58) at (14.75, 2) {};
		\node [text height=1.5 ex, text depth=0.25 ex, style=none] (59) at (-7.25, 1.25) {$W$};
		\node [text height=1.5 ex, text depth=0.25 ex, style=none] (60) at (-4.75, 1.25) {$W$};
		\node [text height=1.5 ex, text depth=0.25 ex, style=none] (61) at (-3.25, 1.25) {$W$};
		\node [text height=1.5 ex, text depth=0.25 ex, style=none] (62) at (-1, 1.25) {$W$};
		\node [text height=1.5 ex, text depth=0.25 ex, style=none, style=none] (63) at (0.25, 1.25) {$W$};
		\node [text height=1.5 ex, text depth=0.25 ex, style=none] (64) at (1.5, 1.25) {$W$};
		\node [text height=1.5 ex, text depth=0.25 ex, style=none] (65) at (4.5, 1.25) {$W$};
		\node [text height=1.5 ex, text depth=0.25 ex, style=none] (66) at (8.5, 1.25) {$W$};
		\node [text height=1.5 ex, text depth=0.25 ex, style=none] (67) at (10.75, 1.25) {$W$};
		\node [text height=1.5 ex, text depth=0.25 ex, style=none] (68) at (12.25, 1.25) {$W$};
		\node [text height=1.5 ex, text depth=0.25 ex, style=none] (69) at (14.75, 1.25) {$W$};
		\node [style=none] (70) at (-7.25, 0.5) {};
		\node [style=none] (71) at (-4.75, 0.5) {};
		\node [style=none] (72) at (-3.25, 0.5) {};
		\node [style=none] (73) at (-1, 0.5) {};
		\node [style=none] (74) at (0.25, 0.5) {};
		\node [style=none] (75) at (1.5, 0.5) {};
		\node [style=none] (76) at (4.5, 0.5) {};
		\node [style=none] (77) at (8.5, 0.5) {};
		\node [style=none] (78) at (10.75, 0.5) {};
		\node [style=none] (79) at (12.25, 0.5) {};
		\node [style=none] (80) at (14.75, 0.5) {};
		\node [draw, circle, minimum size=0.12 cm, fill=white, style=none] (81) at (13.5, -0.25) {};
		\node [style=none] (82) at (0.25, -1.25) {};
		\node [style=none] (83) at (13.5, -1.5) {};
	\end{pgfonlayer}
	\begin{pgfonlayer}{edgelayer}
		\draw [thick] (38.center) to (55.center);
		\draw [thick] (9.center) to (43.center);
		\draw [thick] (20.center) to (47.center);
		\draw [thick] (35.center) to (53.center);
		\draw [thick] (41.center) to (56.center);
		\draw [thick] (22.center) to (8.center);
		\draw [thick] (23.center) to (48.center);
		\draw [thick] (24.center) to (49.center);
		\draw [thick] (44.center) to (46.center);
		\draw [thick] (37.center) to (39.center);
		\draw [thick, bend left=270, looseness=1.25] (77.center) to (78.center);
		\draw [thick] (42.center) to (57.center);
		\draw [thick] (14.center) to (28.center);
		\draw [thick] (30.center) to (51.center);
		\draw [thick] (13.center) to (46.center);
		\draw [thick] (34.center) to (36.center);
		\draw [thick, bend right=90] (79.center) to (80.center);
		\draw [thick] (27.center) to (50.center);
		\draw [thick] (45.center) to (58.center);
		\draw [thick, bend left=270, looseness=1.50] (15.center) to (16.center);
		\draw [thick] (44.center) to (13.center);
		\draw [thick] (10.center) to (21.center);
		\draw [thick] (37.center) to (12.center);
		\draw [thick, looseness=0.00] (18.center) to (29.center);
		\draw [thick, bend left=270, looseness=1.25] (72.center) to (73.center);
		\draw [thick] (8.center) to (25.center);
		\draw [thick, bend left=90, looseness=2.25] (14.center) to (15.center);
		\draw [thick] (34.center) to (11.center);
		\draw [thick] (19.center) to (21.center);
		\draw [thick] (74.center) to (82.center);
		\draw [thick] (11.center) to (36.center);
		\draw [thick] (26.center) to (33.center);
		\draw [thick] (19.center) to (10.center);
		\draw [thick] (7.center) to (33.center);
		\draw [thick] (12.center) to (39.center);
		\draw [thick] (17.center) to (31.center);
		\draw [thick] (22.center) to (25.center);
		\draw [thick] (40.center) to (9.center);
		\draw [thick, bend left=270, looseness=1.25] (70.center) to (71.center);
		\draw [thick, bend left=90, looseness=2.25] (16.center) to (17.center);
		\draw [thick] (26.center) to (7.center);
		\draw [thick] (81.center) to (83.center);
		\draw [thick] (32.center) to (52.center);
		\draw [thick, bend left=270] (75.center) to (76.center);
		\draw [thick] (40.center) to (43.center);
	\end{pgfonlayer}
\end{tikzpicture}

%% file: tikz/into-der2a.tikz
\begin{tikzpicture}
	\begin{pgfonlayer}{nodelayer}
		\node [style=none, text depth=0.25 ex, text height=1.5 ex] (0) at (-8.25, 4.75) {Mary};
		\node [style=none, text depth=0.25 ex, text height=1.5 ex] (1) at (-5.25, 4.75) {likes};
		\node [style=none, text depth=0.25 ex, text height=1.5 ex] (2) at (-2, 4.75) {musicals};
		\node [style=none, text depth=0.25 ex, text height=1.5 ex] (3) at (1.5, 4.75) {Mary};
		\node [style=none, text depth=0.25 ex, text height=1.5 ex] (4) at (4.5, 4.75) {likes};
		\node [style=none, text depth=0.25 ex, text height=1.5 ex] (5) at (7.75, 4.75) {musicals};
		\node [style=none] (6) at (-5.25, 3.5) {};
		\node [style=none] (7) at (4.5, 3.5) {};
		\node [style=none] (8) at (-8.25, 3) {};
		\node [style=none] (9) at (-2, 3) {};
		\node [style=none] (10) at (1.5, 3) {};
		\node [style=none] (11) at (7.75, 3) {};
		\node [style=none] (12) at (-9, 2) {};
		\node [style=none] (13) at (-8.25, 2) {};
		\node [style=none] (14) at (-7.5, 2) {};
		\node [style=none] (15) at (-6.5, 2) {};
		\node [style=none] (16) at (-6, 2) {};
		\node [style=none] (17) at (-4.5, 2) {};
		\node [style=none] (18) at (-4, 2) {};
		\node [style=none] (19) at (-2.75, 2) {};
		\node [style=none] (20) at (-2, 2) {};
		\node [style=none] (21) at (-1.25, 2) {};
		\node [style=none] (22) at (0.75, 2) {};
		\node [style=none] (23) at (1.5, 2) {};
		\node [style=none] (24) at (2.25, 2) {};
		\node [style=none] (25) at (3.25, 2) {};
		\node [style=none] (26) at (3.75, 2) {};
		\node [style=none] (27) at (5.25, 2) {};
		\node [style=none] (28) at (5.75, 2) {};
		\node [style=none] (29) at (7, 2) {};
		\node [style=none] (30) at (7.75, 2) {};
		\node [style=none] (31) at (8.5, 2) {};
		\node [style=none] (32) at (1.5, 1.5) {};
		\node [style=none] (33) at (3.75, 1.5) {};
		\node [style=none] (34) at (5.25, 1.5) {};
		\node [style=none] (35) at (7.75, 1.5) {};
		\node [style=none] (36) at (-8.25, 1.25) {};
		\node [style=none] (37) at (-6, 1.25) {};
		\node [style=none] (38) at (-4.5, 1.25) {};
		\node [style=none] (39) at (-2, 1.25) {};
		\node [style=none] (40) at (-0.25, 0.75) {$=$};
		\node [style=none, text depth=0.25 ex, text height=1.5 ex] (41) at (1.5, 0.75) {$W$};
		\node [style=none, text depth=0.25 ex, text height=1.5 ex] (42) at (3.75, 0.75) {$W$};
		\node [style=none, text depth=0.25 ex, text height=1.5 ex] (43) at (5.25, 0.75) {$W$};
		\node [style=none, text depth=0.25 ex, text height=1.5 ex] (44) at (7.75, 0.75) {$W$};
		\node [style=none, text depth=0.25 ex, text height=1.5 ex] (45) at (-8.25, 0.5) {$W$};
		\node [style=none, text depth=0.25 ex, text height=1.5 ex] (46) at (-6, 0.5) {$W$};
		\node [style=none, text depth=0.25 ex, text height=1.5 ex] (47) at (-4.5, 0.5) {$W$};
		\node [style=none, text depth=0.25 ex, text height=1.5 ex] (48) at (-2, 0.5) {$W$};
		\node [style=none] (49) at (5.25, -0) {};
		\node [style=none] (50) at (7.75, -0) {};
		\node [style=none] (51) at (-8.25, -0.25) {};
		\node [style=none] (52) at (-6, -0.25) {};
		\node [style=none] (53) at (-4.5, -0.25) {};
		\node [style=none] (54) at (-2, -0.25) {};
		\node [style=none, fill=white, minimum size=0.12 cm, circle, draw] (55) at (5.25, -0.5) {};
		\node [style=none, fill=white, minimum size=0.12 cm, circle, draw] (56) at (-3.25, -1.25) {};
		\node [style=none] (57) at (1.5, -1.25) {};
		\node [style=none] (58) at (3.75, -1.25) {};
		\node [style=none] (59) at (4.5, -1.25) {};
		\node [style=none] (60) at (6, -1.25) {};
		\node [style=none] (61) at (6, -1.25) {};
		\node [style=none] (62) at (7.75, -1.25) {};
		\node [style=none] (63) at (4.5, -2) {};
		\node [style=none] (64) at (-3.25, -2.5) {};
		\node [style=none] (65) at (3.75, -0) {};
		\node [style=none] (66) at (1.5, -0) {};
		\node [text height=1.5 ex, text depth=0.25 ex, style=none] (67) at (12.75, 0.75) {$W$};
		\node [style=none] (68) at (13.25, 2) {};
		\node [style=none] (69) at (12.75, 2) {};
		\node [draw, circle, minimum size=0.12 cm, fill=white, style=none] (70) at (12.75, -0.5) {};
		\node [style=none] (71) at (13.5, -1.25) {};
		\node [style=none] (72) at (11.25, 2) {};
		\node [style=none] (73) at (13.5, -1.25) {};
		\node [text height=1.5 ex, text depth=0.25 ex, style=none] (74) at (12, 4.75) {likes};
		\node [style=none] (75) at (12.75, 1.5) {};
		\node [style=none] (76) at (10.75, 2) {};
		\node [text height=1.5 ex, text depth=0.25 ex, style=none] (77) at (11.25, 0.75) {$W$};
		\node [style=none] (78) at (12, -1.25) {};
		\node [style=none] (79) at (11.25, -1.25) {};
		\node [style=none] (80) at (12.75, -0) {};
		\node [style=none] (81) at (11.25, 1.5) {};
		\node [style=none] (82) at (12, 3.5) {};
		\node [style=none] (83) at (11.25, -0) {};
		\node [style=none] (84) at (9.5, 0.75) {$\Rightarrow$};
	\end{pgfonlayer}
	\begin{pgfonlayer}{edgelayer}
		\draw [thick] (19.center) to (21.center);
		\draw [thick] (55.center) to (49.center);
		\draw [thick] (19.center) to (9.center);
		\draw [thick] (11.center) to (31.center);
		\draw [thick] (29.center) to (11.center);
		\draw [thick] (30.center) to (35.center);
		\draw [thick, bend right=90, looseness=1.50] (61.center) to (62.center);
		\draw [thick] (17.center) to (38.center);
		\draw [thick] (23.center) to (32.center);
		\draw [thick] (22.center) to (24.center);
		\draw [thick] (12.center) to (14.center);
		\draw [thick, bend right=90, looseness=1.25] (53.center) to (54.center);
		\draw [thick] (10.center) to (24.center);
		\draw [thick] (7.center) to (28.center);
		\draw [thick] (25.center) to (28.center);
		\draw [thick] (20.center) to (39.center);
		\draw [thick] (59.center) to (63.center);
		\draw [thick, bend right=90, looseness=1.25] (57.center) to (58.center);
		\draw [thick] (22.center) to (10.center);
		\draw [thick] (16.center) to (37.center);
		\draw [thick] (25.center) to (7.center);
		\draw [thick] (15.center) to (6.center);
		\draw [thick] (15.center) to (18.center);
		\draw [thick] (26.center) to (33.center);
		\draw [thick] (12.center) to (8.center);
		\draw [thick] (29.center) to (31.center);
		\draw [thick] (6.center) to (18.center);
		\draw [thick] (8.center) to (14.center);
		\draw [thick] (27.center) to (34.center);
		\draw [thick] (56.center) to (64.center);
		\draw [thick] (9.center) to (21.center);
		\draw [thick] (50.center) to (62.center);
		\draw [thick, bend left=90, looseness=1.75] (59.center) to (60.center);
		\draw [thick] (13.center) to (36.center);
		\draw [thick, bend right=90, looseness=1.25] (51.center) to (52.center);
		\draw [thick] (65.center) to (58.center);
		\draw [thick] (66.center) to (57.center);
		\draw [thick] (70.center) to (80.center);
		\draw [thick] (82.center) to (68.center);
		\draw [thick] (76.center) to (68.center);
		\draw [thick] (76.center) to (82.center);
		\draw [thick] (72.center) to (81.center);
		\draw [thick] (69.center) to (75.center);
		\draw [thick, bend left=90, looseness=1.75] (78.center) to (73.center);
		\draw [thick] (83.center) to (79.center);
	\end{pgfonlayer}
\end{tikzpicture}

%% file: tikz/tworhemes-preg.tikz
\begin{tikzpicture}
	\begin{pgfonlayer}{nodelayer}
		\node [text height=1.5 ex, text depth=0.25 ex, style=none] (0) at (-11.25, 2.5) {John};
		\node [style=none] (1) at (-8, 2.5) {$\triangleleft$};
		\node [text height=1.5 ex, text depth=0.25 ex, style=none] (2) at (-4.5, 2.5) {likes};
		\node [style=none] (3) at (-1, 2.5) {$\triangleright$};
		\node [text height=1.5 ex, text depth=0.25 ex, style=none] (4) at (2.25, 2.5) {Mary};
		\node [text height=1.5 ex, text depth=0.25 ex, style=none] (5) at (-11.25, 1.25) {$\rho$};
		\node [text height=1.5 ex, text depth=0.25 ex, style=none] (6) at (-9, 1.25) {$\rho^r$};
		\node [text height=1.5 ex, text depth=0.25 ex, style=none, style=none] (7) at (-8, 1.25) {$s$};
		\node [text height=1.5 ex, text depth=0.25 ex, style=none] (8) at (-7, 1.25) {$\theta^l$};
		\node [text height=1.5 ex, text depth=0.25 ex, style=none] (9) at (-5, 1.25) {$\theta$};
		\node [text height=1.5 ex, text depth=0.25 ex, style=none] (10) at (-4, 1.25) {$\theta$};
		\node [text height=1.5 ex, text depth=0.25 ex, style=none] (11) at (-2, 1.25) {$\theta^r$};
		\node [text height=1.5 ex, text depth=0.25 ex, style=none, style=none] (12) at (-1, 1.25) {$s$};
		\node [text height=1.5 ex, text depth=0.25 ex, style=none] (13) at (0, 1.25) {$\rho^l$};
		\node [text height=1.5 ex, text depth=0.25 ex, style=none] (14) at (2.25, 1.25) {$\rho$};
		\node [style=none] (15) at (-11.25, 0.5) {};
		\node [style=none] (16) at (-9, 0.5) {};
		\node [style=none] (17) at (-9, 0.5) {};
		\node [style=none] (18) at (-8, 0.5) {};
		\node [style=none] (19) at (-7, 0.5) {};
		\node [style=none] (20) at (-7, 0.5) {};
		\node [style=none] (21) at (-5, 0.5) {};
		\node [style=none] (22) at (-4, 0.5) {};
		\node [style=none] (23) at (-2, 0.5) {};
		\node [style=none] (24) at (-1, 0.5) {};
		\node [style=none] (25) at (0, 0.5) {};
		\node [style=none] (26) at (2.25, 0.5) {};
		\node [style=none] (27) at (-8, -1.25) {};
		\node [style=none] (28) at (-1, -1.25) {};
	\end{pgfonlayer}
	\begin{pgfonlayer}{edgelayer}
		\draw [thick, bend left=270, looseness=1.25] (15.center) to (17.center);
		\draw [thick, bend left=270, looseness=1.25] (19.center) to (21.center);
		\draw [thick] (24.center) to (28.center);
		\draw [thick, bend left=270, looseness=1.25] (25.center) to (26.center);
		\draw [thick] (18.center) to (27.center);
		\draw [thick, bend left=270, looseness=1.25] (22.center) to (23.center);
	\end{pgfonlayer}
\end{tikzpicture}

%% file: tikz/tworhemes1.tikz
\begin{tikzpicture}
	\begin{pgfonlayer}{nodelayer}
		\node [style=none] (0) at (0, 5.5) {$\triangleright$};
		\node [style=none, text depth=0.25 ex, text height=1.5 ex] (1) at (-10, 5.25) {John};
		\node [style=none, text depth=0.25 ex, text height=1.5 ex] (2) at (-3.5, 5.25) {likes};
		\node [style=none, text depth=0.25 ex, text height=1.5 ex] (3) at (2.75, 5.25) {Mary};
		\node [style=none, text depth=0.25 ex, text height=1.5 ex] (4) at (5.75, 5.5) {John};
		\node [style=none, text depth=0.25 ex, text height=1.5 ex] (5) at (8.25, 5.5) {likes};
		\node [style=none, text depth=0.25 ex, text height=1.5 ex] (6) at (11, 5.5) {Mary};
		\node [style=none] (7) at (0, 4.75) {};
		\node [style=none] (8) at (-3.5, 4.25) {};
		\node [style=none] (9) at (8.25, 4.5) {};
		\node [style=none] (10) at (-10, 3.75) {};
		\node [style=none] (11) at (3, 3.75) {};
		\node [style=none] (12) at (5.5, 4) {};
		\node [style=none] (13) at (11, 4) {};
		\node [style=none] (14) at (-1.25, 3.5) {};
		\node [style=none] (15) at (-0.5, 3.5) {};
		\node [style=none] (16) at (0.5, 3.5) {};
		\node [style=none] (17) at (1.25, 3.5) {};
		\node [style=none, fill=white, minimum size=0.12 cm, circle, draw] (18) at (0, 3) {};
		\node [style=none] (19) at (-10.75, 2.75) {};
		\node [style=none] (20) at (-10, 2.75) {};
		\node [style=none] (21) at (-9.25, 2.75) {};
		\node [style=none] (22) at (-4.75, 2.75) {};
		\node [style=none] (23) at (-4.25, 2.75) {};
		\node [style=none] (24) at (-2.75, 2.75) {};
		\node [style=none] (25) at (-2.25, 2.75) {};
		\node [style=none] (26) at (-2, 2.75) {};
		\node [style=none] (27) at (-1.25, 2.75) {};
		\node [style=none] (28) at (-1.25, 2.75) {};
		\node [style=none] (29) at (0, 2.75) {};
		\node [style=none] (30) at (0, 2.75) {};
		\node [style=none] (31) at (1.25, 2.75) {};
		\node [style=none] (32) at (1.25, 2.75) {};
		\node [style=none] (33) at (2, 2.75) {};
		\node [style=none] (34) at (2.25, 2.75) {};
		\node [style=none] (35) at (3, 2.75) {};
		\node [style=none] (36) at (3.75, 2.75) {};
		\node [style=none] (37) at (4.75, 3) {};
		\node [style=none] (38) at (5.5, 3) {};
		\node [style=none] (39) at (6.25, 3) {};
		\node [style=none] (40) at (7, 3) {};
		\node [style=none] (41) at (7.5, 3) {};
		\node [style=none] (42) at (9, 3) {};
		\node [style=none] (43) at (9.5, 3) {};
		\node [style=none] (44) at (10.25, 3) {};
		\node [style=none] (45) at (11, 3) {};
		\node [style=none] (46) at (11.75, 3) {};
		\node [style=none] (47) at (-10, 1.75) {};
		\node [style=none] (48) at (-4.25, 1.75) {};
		\node [style=none] (49) at (-2.75, 1.75) {};
		\node [style=none] (50) at (-1.25, 1.75) {};
		\node [style=none] (51) at (0, 1.5) {};
		\node [style=none] (52) at (1.25, 1.75) {};
		\node [style=none] (53) at (3, 1.75) {};
		\node [style=none] (54) at (4.25, 2) {$=$};
		\node [style=none] (55) at (5.5, 2.25) {};
		\node [style=none] (56) at (7.5, 2.25) {};
		\node [style=none] (57) at (9, 2.25) {};
		\node [style=none] (58) at (11, 2.25) {};
		\node [style=none] (59) at (-2.75, 1.75) {};
		\node [style=none] (60) at (-1.25, 1.75) {};
		\node [style=none] (61) at (0, 1.5) {};
		\node [style=none] (62) at (1.25, 1.75) {};
		\node [style=none] (63) at (3, 1.75) {};
		\node [style=none] (64) at (5.5, 2.25) {};
		\node [style=none] (65) at (7.5, 2.25) {};
		\node [style=none] (66) at (9, 2.25) {};
		\node [style=none] (67) at (11, 2.25) {};
		\node [style=none, fill=white, minimum size=0.12 cm, circle, draw] (68) at (10, 1.5) {};
		\node [style=none] (69) at (0, 0.5) {};
		\node [style=none] (70) at (10, 0.5) {};
		\node [style=none] (71) at (-7, 5.5) {$\triangleleft$};
		\node [style=none] (72) at (-5.75, 1.75) {};
		\node [style=none] (73) at (-5, 2.75) {};
		\node [style=none] (74) at (-7, 1.5) {};
		\node [style=none] (75) at (-7, 0.5) {};
		\node [style=none] (76) at (-5.75, 2.75) {};
		\node [style=none] (77) at (-8.25, 1.75) {};
		\node [style=none] (78) at (-5.75, 3.5) {};
		\node [style=none] (79) at (-7, 1.5) {};
		\node [draw, circle, minimum size=0.12 cm, fill=white, style=none] (80) at (-7, 3) {};
		\node [style=none] (81) at (-8.25, 3.5) {};
		\node [style=none] (82) at (-5.75, 2.75) {};
		\node [style=none] (83) at (-8.25, 2.75) {};
		\node [style=none] (84) at (-7.5, 3.5) {};
		\node [style=none] (85) at (-6.5, 3.5) {};
		\node [style=none] (86) at (-7, 4.75) {};
		\node [style=none] (87) at (-8.25, 2.75) {};
		\node [style=none] (88) at (-7, 2.75) {};
		\node [style=none] (89) at (-9, 2.75) {};
		\node [style=none] (90) at (-7, 2.75) {};
		\node [style=none] (91) at (6.5, 0.5) {};
		\node [draw, circle, minimum size=0.12 cm, fill=white, style=none] (92) at (6.5, 1.5) {};
		\node [style=none] (93) at (-8.25, 1.75) {};
		\node [style=none] (94) at (-10, 1.75) {};
		\node [style=none] (95) at (-4.25, 1.75) {};
		\node [style=none] (96) at (-5.75, 1.75) {};
	\end{pgfonlayer}
	\begin{pgfonlayer}{edgelayer}
		\draw [thick] (38.center) to (55.center);
		\draw [thick] (9.center) to (43.center);
		\draw [thick] (20.center) to (47.center);
		\draw [thick] (35.center) to (53.center);
		\draw [thick] (41.center) to (56.center);
		\draw [thick] (22.center) to (8.center);
		\draw [thick] (23.center) to (48.center);
		\draw [thick] (24.center) to (49.center);
		\draw [thick] (44.center) to (46.center);
		\draw [thick] (37.center) to (39.center);
		\draw [thick, bend right=90, looseness=1.25] (64.center) to (65.center);
		\draw [thick] (42.center) to (57.center);
		\draw [thick] (14.center) to (28.center);
		\draw [thick] (30.center) to (51.center);
		\draw [thick] (13.center) to (46.center);
		\draw [thick] (34.center) to (36.center);
		\draw [thick, bend right=90, looseness=1.25] (66.center) to (67.center);
		\draw [thick] (27.center) to (50.center);
		\draw [thick] (45.center) to (58.center);
		\draw [thick, bend right=90, looseness=1.50] (15.center) to (16.center);
		\draw [thick] (44.center) to (13.center);
		\draw [thick] (10.center) to (21.center);
		\draw [thick] (37.center) to (12.center);
		\draw [thick] (18.center) to (29.center);
		\draw [thick, bend right=90, looseness=1.50] (59.center) to (60.center);
		\draw [thick] (8.center) to (25.center);
		\draw [thick, bend left=90, looseness=2.25] (14.center) to (15.center);
		\draw [thick] (34.center) to (11.center);
		\draw [thick] (19.center) to (21.center);
		\draw [thick] (61.center) to (69.center);
		\draw [thick] (11.center) to (36.center);
		\draw [thick] (26.center) to (33.center);
		\draw [thick] (19.center) to (10.center);
		\draw [thick] (7.center) to (33.center);
		\draw [thick] (12.center) to (39.center);
		\draw [thick] (17.center) to (31.center);
		\draw [thick] (22.center) to (25.center);
		\draw [thick] (40.center) to (9.center);
		\draw [thick, bend left=90, looseness=2.25] (16.center) to (17.center);
		\draw [thick] (26.center) to (7.center);
		\draw [thick] (68.center) to (70.center);
		\draw [thick] (32.center) to (52.center);
		\draw [thick, bend right=90, looseness=1.25] (62.center) to (63.center);
		\draw [thick] (40.center) to (43.center);
		\draw [thick] (81.center) to (83.center);
		\draw [thick] (88.center) to (74.center);
		\draw [thick] (87.center) to (77.center);
		\draw [thick, bend right=90, looseness=1.50] (84.center) to (85.center);
		\draw [thick] (80.center) to (90.center);
		\draw [thick, bend left=90, looseness=2.25] (81.center) to (84.center);
		\draw [thick] (79.center) to (75.center);
		\draw [thick] (89.center) to (73.center);
		\draw [thick] (86.center) to (73.center);
		\draw [thick] (78.center) to (82.center);
		\draw [thick, bend left=90, looseness=2.25] (85.center) to (78.center);
		\draw [thick] (89.center) to (86.center);
		\draw [thick] (76.center) to (72.center);
		\draw [thick] (92.center) to (91.center);
		\draw [thick, bend right=90, looseness=1.25] (94.center) to (93.center);
		\draw [thick, bend right=90, looseness=1.50] (96.center) to (95.center);
	\end{pgfonlayer}
\end{tikzpicture}

%% file: tikz/rheme-verb.tikz
\begin{tikzpicture}
	\begin{pgfonlayer}{nodelayer}
		\node [text height=1.5 ex, text depth=0.25 ex, style=none] (0) at (-11.25, 2.5) {John};
		\node [style=none] (1) at (-8, 2.5) {$\triangleright$};
		\node [text height=1.5 ex, text depth=0.25 ex, style=none] (2) at (-4.5, 2.5) {likes};
		\node [style=none] (3) at (-1, 2.5) {$\triangleleft$};
		\node [text height=1.5 ex, text depth=0.25 ex, style=none] (4) at (2.25, 2.5) {Mary};
		\node [text height=1.5 ex, text depth=0.25 ex, style=none] (5) at (-11.25, 1.25) {$\theta$};
		\node [text height=1.5 ex, text depth=0.25 ex, style=none] (6) at (-9, 1.25) {$\theta^r$};
		\node [text height=1.5 ex, text depth=0.25 ex, style=none, style=none] (7) at (-8, 1.25) {$s$};
		\node [text height=1.5 ex, text depth=0.25 ex, style=none] (8) at (-7, 1.25) {$\rho^l$};
		\node [text height=1.5 ex, text depth=0.25 ex, style=none] (9) at (-5, 1.25) {$\rho$};
		\node [text height=1.5 ex, text depth=0.25 ex, style=none] (10) at (-4, 1.25) {$\rho$};
		\node [text height=1.5 ex, text depth=0.25 ex, style=none] (11) at (-2, 1.25) {$\rho^r$};
		\node [text height=1.5 ex, text depth=0.25 ex, style=none, style=none] (12) at (-1, 1.25) {$s$};
		\node [text height=1.5 ex, text depth=0.25 ex, style=none] (13) at (0, 1.25) {$\theta^l$};
		\node [text height=1.5 ex, text depth=0.25 ex, style=none] (14) at (2.25, 1.25) {$\theta$};
		\node [style=none] (15) at (-11.25, 0.5) {};
		\node [style=none] (16) at (-9, 0.5) {};
		\node [style=none] (17) at (-9, 0.5) {};
		\node [style=none] (18) at (-8, 0.5) {};
		\node [style=none] (19) at (-7, 0.5) {};
		\node [style=none] (20) at (-7, 0.5) {};
		\node [style=none] (21) at (-5, 0.5) {};
		\node [style=none] (22) at (-4, 0.5) {};
		\node [style=none] (23) at (-2, 0.5) {};
		\node [style=none] (24) at (-1, 0.5) {};
		\node [style=none] (25) at (0, 0.5) {};
		\node [style=none] (26) at (2.25, 0.5) {};
		\node [style=none] (27) at (-8, -1.25) {};
		\node [style=none] (28) at (-1, -1.25) {};
	\end{pgfonlayer}
	\begin{pgfonlayer}{edgelayer}
		\draw [thick, bend left=270, looseness=1.25] (22.center) to (23.center);
		\draw [thick, bend left=270, looseness=1.25] (20.center) to (21.center);
		\draw [thick, bend left=270, looseness=1.25] (15.center) to (17.center);
		\draw [thick] (24.center) to (28.center);
		\draw [thick] (18.center) to (27.center);
		\draw [thick, bend left=270, looseness=1.25] (25.center) to (26.center);
	\end{pgfonlayer}
\end{tikzpicture}

%% file: tikz/nested-normal.tikz
\begin{tikzpicture}
	\begin{pgfonlayer}{nodelayer}
		\node [text height=1.5 ex, text depth=0.25 ex, style=none] (0) at (-11.25, 2.5) {Mary};
		\node [text height=1.5 ex, text depth=0.25 ex, style=none] (1) at (-8.25, 2.5) {wrote};
		\node [text height=1.5 ex, text depth=0.25 ex, style=none] (2) at (-5, 2.5) {a book};
		\node [text height=1.5 ex, text depth=0.25 ex, style=none] (3) at (-2, 2.5) {about};
		\node [text height=1.5 ex, text depth=0.25 ex, style=none] (4) at (1.25, 2.5) {art};
		\node [text height=1.5 ex, text depth=0.25 ex, style=none] (5) at (-11.25, 1.25) {$n$};
		\node [text height=1.5 ex, text depth=0.25 ex, style=none] (6) at (-9, 1.25) {$n^r$};
		\node [text height=1.5 ex, text depth=0.25 ex, style=none, style=none] (7) at (-8, 1.25) {$s$};
		\node [text height=1.5 ex, text depth=0.25 ex, style=none] (8) at (-7.25, 1.25) {$n^l$};
		\node [text height=1.5 ex, text depth=0.25 ex, style=none] (9) at (-5, 1.25) {$n$};
		\node [text height=1.5 ex, text depth=0.25 ex, style=none] (10) at (-2.75, 1.25) {$n^l$};
		\node [text height=1.5 ex, text depth=0.25 ex, style=none] (11) at (-2, 1.25) {$n$};
		\node [text height=1.5 ex, text depth=0.25 ex, style=none] (12) at (-1, 1.25) {$n^r$};
		\node [text height=1.5 ex, text depth=0.25 ex, style=none] (13) at (1.25, 1.25) {$n$};
		\node [style=none] (14) at (-11.25, 0.5) {};
		\node [style=none] (15) at (-9, 0.5) {};
		\node [style=none] (16) at (-9, 0.5) {};
		\node [style=none] (17) at (-8, 0.5) {};
		\node [style=none] (18) at (-7.25, 0.5) {};
		\node [style=none] (19) at (-5, 0.5) {};
		\node [style=none] (20) at (-2.75, 0.5) {};
		\node [style=none] (21) at (-2, 0.5) {};
		\node [style=none] (22) at (-1, 0.5) {};
		\node [style=none] (23) at (1.25, 0.5) {};
		\node [style=none] (24) at (-8, -1.25) {};
	\end{pgfonlayer}
	\begin{pgfonlayer}{edgelayer}
		\draw [thick, bend left=270, looseness=1.25] (22.center) to (23.center);
		\draw [thick, bend left=270, looseness=1.25] (14.center) to (15.center);
		\draw [thick] (17.center) to (24.center);
		\draw [thick, bend left=270, looseness=1.25] (19.center) to (20.center);
		\draw [thick, bend right=90] (18.center) to (21.center);
	\end{pgfonlayer}
\end{tikzpicture}

%% file: tikz/nested.tikz
\begin{tikzpicture}
	\begin{pgfonlayer}{nodelayer}
		\node [text height=1.5 ex, text depth=0.25 ex, style=none] (0) at (-11.25, 2.5) {Mary};
		\node [text height=1.5 ex, text depth=0.25 ex, style=none] (1) at (-8, 2.5) {wrote};
		\node [text height=1.5 ex, text depth=0.25 ex, style=none] (2) at (-4.5, 2.5) {a book};
		\node [text height=1.5 ex, text depth=0.25 ex, style=none] (3) at (-1.25, 2.5) {about};
		\node [text height=1.5 ex, text depth=0.25 ex, style=none] (4) at (2.25, 2.5) {$\triangleright$};
		\node [text height=1.5 ex, text depth=0.25 ex, style=none] (5) at (5.5, 2.5) {art};
		\node [text height=1.5 ex, text depth=0.25 ex, style=none] (6) at (-11.25, 1.25) {$n$};
		\node [text height=1.5 ex, text depth=0.25 ex, style=none] (7) at (-9, 1.25) {$n^r$};
		\node [text height=1.5 ex, text depth=0.25 ex, style=none, style=none] (8) at (-8, 1.25) {$\theta$};
		\node [text height=1.5 ex, text depth=0.25 ex, style=none] (9) at (-7, 1.25) {$n^l$};
		\node [text height=1.5 ex, text depth=0.25 ex, style=none] (10) at (-4.5, 1.25) {$n$};
		\node [text height=1.5 ex, text depth=0.25 ex, style=none] (11) at (-1.75, 1.25) {$n^r$};
		\node [text height=1.5 ex, text depth=0.25 ex, style=none] (12) at (-0.75, 1.25) {$n$};
		\node [text height=1.5 ex, text depth=0.25 ex, style=none] (13) at (1.25, 1.25) {$\theta^r$};
		\node [text height=1.5 ex, text depth=0.25 ex, style=none, style=none] (14) at (2.25, 1.25) {$s$};
		\node [text height=1.5 ex, text depth=0.25 ex, style=none] (15) at (3.25, 1.25) {$\rho^l$};
		\node [text height=1.5 ex, text depth=0.25 ex, style=none] (16) at (5.5, 1.25) {$\rho$};
		\node [style=none] (17) at (-11.25, 0.5) {};
		\node [style=none] (18) at (-9, 0.5) {};
		\node [style=none] (19) at (-9, 0.5) {};
		\node [style=none] (20) at (-8, 0.5) {};
		\node [style=none] (21) at (-7, 0.5) {};
		\node [style=none] (22) at (-4.5, 0.5) {};
		\node [style=none] (23) at (-4.5, 0.5) {};
		\node [style=none] (24) at (-1.75, 0.5) {};
		\node [style=none] (25) at (-0.75, 0.5) {};
		\node [style=none] (26) at (1.25, 0.5) {};
		\node [style=none] (27) at (2.25, 0.5) {};
		\node [style=none] (28) at (3.25, 0.5) {};
		\node [style=none] (29) at (5.5, 0.5) {};
		\node [style=none] (30) at (2.25, -1.25) {};
	\end{pgfonlayer}
	\begin{pgfonlayer}{edgelayer}
		\draw [thick, bend left=270, looseness=0.75] (21.center) to (25.center);
		\draw [thick, bend left=270, looseness=0.75] (20.center) to (26.center);
		\draw [thick, bend left=270, looseness=1.25] (17.center) to (19.center);
		\draw [thick] (27.center) to (30.center);
		\draw [thick, bend left=270, looseness=1.25] (22.center) to (24.center);
		\draw [thick, bend left=270, looseness=1.25] (28.center) to (29.center);
	\end{pgfonlayer}
\end{tikzpicture}

%% file: tikz/nested1.tikz
\begin{tikzpicture}
	\begin{pgfonlayer}{nodelayer}
		\node [text height=1.5 ex, text depth=0.25 ex, style=none] (0) at (-11.25, 2.5) {Mary};
		\node [text height=1.5 ex, text depth=0.25 ex, style=none] (1) at (-8.5, 2.5) {wrote};
		\node [text height=1.5 ex, text depth=0.25 ex, style=none] (2) at (-4.5, 2.5) {$\triangleright$};
		\node [text height=1.5 ex, text depth=0.25 ex, style=none] (3) at (-1.25, 2.5) {a book};
		\node [text height=1.5 ex, text depth=0.25 ex, style=none] (4) at (2, 2.5) {$\triangleleft$};
		\node [text height=1.5 ex, text depth=0.25 ex, style=none] (5) at (6, 2.5) {about};
		\node [text height=1.5 ex, text depth=0.25 ex, style=none] (6) at (8.75, 2.5) {art};
		\node [text height=1.5 ex, text depth=0.25 ex, style=none] (7) at (-11.25, 1.25) {$n$};
		\node [text height=1.5 ex, text depth=0.25 ex, style=none] (8) at (-9, 1.25) {$n^r$};
		\node [text height=1.5 ex, text depth=0.25 ex, style=none, style=none] (9) at (-8, 1.25) {$\theta$};
		\node [text height=1.5 ex, text depth=0.25 ex, style=none] (10) at (-5.5, 1.25) {$\theta^r$};
		\node [text height=1.5 ex, text depth=0.25 ex, style=none, style=none] (11) at (-4.5, 1.25) {$\rho$};
		\node [text height=1.5 ex, text depth=0.25 ex, style=none] (12) at (-3.5, 1.25) {$\rho^l$};
		\node [text height=1.5 ex, text depth=0.25 ex, style=none] (13) at (-1.25, 1.25) {$\rho$};
		\node [text height=1.5 ex, text depth=0.25 ex, style=none] (14) at (1, 1.25) {$\rho^r$};
		\node [text height=1.5 ex, text depth=0.25 ex, style=none, style=none] (15) at (2, 1.25) {$s$};
		\node [text height=1.5 ex, text depth=0.25 ex, style=none] (16) at (3, 1.25) {$\theta^l$};
		\node [text height=1.5 ex, text depth=0.25 ex, style=none] (17) at (5.5, 1.25) {$\theta$};
		\node [text height=1.5 ex, text depth=0.25 ex, style=none] (18) at (6.5, 1.25) {$n^r$};
		\node [text height=1.5 ex, text depth=0.25 ex, style=none] (19) at (8.75, 1.25) {$n$};
		\node [style=none] (20) at (-11.25, 0.5) {};
		\node [style=none] (21) at (-9, 0.5) {};
		\node [style=none] (22) at (-9, 0.5) {};
		\node [style=none] (23) at (-8, 0.5) {};
		\node [style=none] (24) at (-8, 0.5) {};
		\node [style=none] (25) at (-5.5, 0.5) {};
		\node [style=none] (26) at (-4.5, 0.5) {};
		\node [style=none] (27) at (-3.5, 0.5) {};
		\node [style=none] (28) at (-1.25, 0.5) {};
		\node [style=none] (29) at (1, 0.5) {};
		\node [style=none] (30) at (2, 0.5) {};
		\node [style=none] (31) at (3, 0.5) {};
		\node [style=none] (32) at (5.5, 0.5) {};
		\node [style=none] (33) at (6.5, 0.5) {};
		\node [style=none] (34) at (8.75, 0.5) {};
		\node [style=none] (35) at (2, -1.25) {};
	\end{pgfonlayer}
	\begin{pgfonlayer}{edgelayer}
		\draw [thick, bend left=270, looseness=1.25] (33.center) to (34.center);
		\draw [thick, bend left=270, looseness=1.25] (23.center) to (25.center);
		\draw [thick, bend left=270, looseness=1.25] (20.center) to (21.center);
		\draw [thick, bend right=90] (26.center) to (29.center);
		\draw [thick, bend left=270, looseness=1.25] (27.center) to (28.center);
		\draw [thick, bend left=270, looseness=1.25] (31.center) to (32.center);
		\draw [thick] (30.center) to (35.center);
	\end{pgfonlayer}
\end{tikzpicture}

%% file: tikz/middle.tikz
\begin{tikzpicture}
	\begin{pgfonlayer}{nodelayer}
		\node [style=none, text depth=0.25 ex, text height=1.5 ex] (0) at (-7.25, 4.75) {Mary};
		\node [style=none, text depth=0.25 ex, text height=1.5 ex] (1) at (-4.25, 4.75) {wrote};
		\node [style=none, text depth=0.25 ex, text height=1.5 ex] (2) at (-1, 4.75) {a book};
		\node [style=none, text depth=0.25 ex, text height=1.5 ex] (3) at (2, 4.75) {about};
		\node [style=none, text depth=0.25 ex, text height=1.5 ex] (4) at (5.25, 4.75) {art};
		\node [style=none] (5) at (-4.25, 3.5) {};
		\node [style=none] (6) at (2, 3.5) {};
		\node [style=none] (7) at (-7.25, 3) {};
		\node [style=none] (8) at (-1, 3) {};
		\node [style=none] (9) at (5.25, 3) {};
		\node [style=none] (10) at (-8, 2) {};
		\node [style=none] (11) at (-7.25, 2) {};
		\node [style=none] (12) at (-6.5, 2) {};
		\node [style=none] (13) at (-5.5, 2) {};
		\node [style=none] (14) at (-5, 2) {};
		\node [style=none] (15) at (-3.5, 2) {};
		\node [style=none] (16) at (-3, 2) {};
		\node [style=none] (17) at (-1.75, 2) {};
		\node [style=none] (18) at (-1, 2) {};
		\node [style=none] (19) at (-0.25, 2) {};
		\node [style=none] (20) at (0.75, 2) {};
		\node [style=none] (21) at (1.25, 2) {};
		\node [style=none] (22) at (2.75, 2) {};
		\node [style=none] (23) at (3.25, 2) {};
		\node [style=none] (24) at (4.5, 2) {};
		\node [style=none] (25) at (5.25, 2) {};
		\node [style=none] (26) at (6, 2) {};
		\node [style=none] (27) at (-7.25, 1.25) {};
		\node [style=none] (28) at (-7.25, 1.25) {};
		\node [style=none] (29) at (-5, 1.25) {};
		\node [style=none] (30) at (-5, 1.25) {};
		\node [style=none] (31) at (-3.5, 1.25) {};
		\node [style=none] (32) at (-3.5, 1.25) {};
		\node [style=none] (33) at (-1, 1.25) {};
		\node [style=none] (34) at (-1, 1.25) {};
		\node [style=none] (35) at (2.75, 1.25) {};
		\node [style=none] (36) at (2.75, 1.25) {};
		\node [style=none] (37) at (5.25, 1.25) {};
		\node [style=none] (38) at (5.25, 1.25) {};
		\node [style=none, fill=white, minimum size=0.12 cm, circle, draw] (39) at (-2.25, 0.25) {};
		\node [style=none] (40) at (-2.25, -0) {};
		\node [style=none] (41) at (1.25, -0) {};
		\node [style=none] (42) at (1.25, -0) {};
		\node [style=none, fill=white, minimum size=0.12 cm, circle, draw] (43) at (-0.5, -1) {};
		\node [style=none] (44) at (-0.5, -2) {};
		\node [style=none] (45) at (-0.5, -2) {};
		\node [style=none] (46) at (-0.5, -2) {};
	\end{pgfonlayer}
	\begin{pgfonlayer}{edgelayer}
		\draw [thick] (7.center) to (12.center);
		\draw [thick] (14.center) to (30.center);
		\draw [thick] (21.center) to (42.center);
		\draw [thick] (13.center) to (16.center);
		\draw [thick] (17.center) to (19.center);
		\draw [thick] (43.center) to (46.center);
		\draw [thick] (8.center) to (19.center);
		\draw [thick] (13.center) to (5.center);
		\draw [thick, bend right=90, looseness=1.25] (36.center) to (38.center);
		\draw [thick] (18.center) to (33.center);
		\draw [thick, bend right=90, looseness=1.25] (31.center) to (34.center);
		\draw [thick] (6.center) to (23.center);
		\draw [thick] (25.center) to (37.center);
		\draw [thick] (17.center) to (8.center);
		\draw [thick] (15.center) to (32.center);
		\draw [thick] (22.center) to (35.center);
		\draw [thick] (10.center) to (12.center);
		\draw [thick] (20.center) to (23.center);
		\draw [thick] (24.center) to (9.center);
		\draw [thick, bend right=90, looseness=1.50] (28.center) to (29.center);
		\draw [thick] (5.center) to (16.center);
		\draw [thick, bend right=90, looseness=1.00] (40.center) to (41.center);
		\draw [thick] (10.center) to (7.center);
		\draw [thick] (11.center) to (27.center);
		\draw [thick] (9.center) to (26.center);
		\draw [thick] (20.center) to (6.center);
		\draw [thick] (24.center) to (26.center);
		\draw [thick] (40.center) to (39.center);
	\end{pgfonlayer}
\end{tikzpicture}

%% file: tikz/middle-normal.tikz
\begin{tikzpicture}
	\begin{pgfonlayer}{nodelayer}
		\node [style=none, text depth=0.25 ex, text height=1.5 ex] (0) at (-7.25, 4.75) {Mary};
		\node [style=none, text depth=0.25 ex, text height=1.5 ex] (1) at (-4.25, 4.75) {wrote};
		\node [style=none, text depth=0.25 ex, text height=1.5 ex] (2) at (-1, 4.75) {a book};
		\node [style=none, text depth=0.25 ex, text height=1.5 ex] (3) at (2.25, 4.75) {about};
		\node [style=none, text depth=0.25 ex, text height=1.5 ex] (4) at (5.5, 4.75) {art};
		\node [style=none] (5) at (-4.25, 3.5) {};
		\node [style=none] (6) at (2.25, 3.5) {};
		\node [style=none] (7) at (-7.25, 3) {};
		\node [style=none] (8) at (-1, 3) {};
		\node [style=none] (9) at (5.5, 3) {};
		\node [style=none] (10) at (-8, 2) {};
		\node [style=none] (11) at (-7.25, 2) {};
		\node [style=none] (12) at (-6.5, 2) {};
		\node [style=none] (13) at (-5.5, 2) {};
		\node [style=none] (14) at (-5, 2) {};
		\node [style=none] (15) at (-3.5, 2) {};
		\node [style=none] (16) at (-3, 2) {};
		\node [style=none] (17) at (-1.75, 2) {};
		\node [style=none] (18) at (-1, 2) {};
		\node [style=none] (19) at (-0.25, 2) {};
		\node [style=none] (20) at (1, 2) {};
		\node [style=none] (21) at (1.5, 2) {};
		\node [style=none] (22) at (3, 2) {};
		\node [style=none] (23) at (3.5, 2) {};
		\node [style=none] (24) at (4.75, 2) {};
		\node [style=none] (25) at (5.5, 2) {};
		\node [style=none] (26) at (6.25, 2) {};
		\node [style=none] (27) at (-7.25, 1.25) {};
		\node [style=none] (28) at (-7.25, 1.25) {};
		\node [style=none] (29) at (-5, 1.25) {};
		\node [style=none] (30) at (-5, 1.25) {};
		\node [style=none] (31) at (-3.5, 1.25) {};
		\node [style=none] (32) at (-3.5, 1.25) {};
		\node [style=none] (33) at (-3.5, 1.25) {};
		\node [style=none] (34) at (1.5, 1.25) {};
		\node [style=none] (35) at (1.5, 1.25) {};
		\node [style=none] (36) at (3, 1.25) {};
		\node [style=none] (37) at (3, 1.25) {};
		\node [style=none] (38) at (5.5, 1.25) {};
		\node [style=none] (39) at (5.5, 1.25) {};
		\node [style=none, fill=white, minimum size=0.12 cm, circle, draw] (40) at (-1, -0.25) {};
		\node [style=none] (41) at (-1, -0.25) {};
		\node [style=none] (42) at (-1, -1.25) {};
	\end{pgfonlayer}
	\begin{pgfonlayer}{edgelayer}
		\draw [thick] (7.center) to (12.center);
		\draw [thick] (14.center) to (29.center);
		\draw [thick] (21.center) to (35.center);
		\draw [thick] (13.center) to (16.center);
		\draw [thick] (17.center) to (19.center);
		\draw [thick] (40.center) to (42.center);
		\draw [thick] (8.center) to (19.center);
		\draw [thick] (13.center) to (5.center);
		\draw [thick, bend right=90, looseness=1.25] (36.center) to (39.center);
		\draw [thick] (18.center) to (41.center);
		\draw [thick] (6.center) to (23.center);
		\draw [thick] (25.center) to (38.center);
		\draw [thick] (17.center) to (8.center);
		\draw [thick] (15.center) to (33.center);
		\draw [thick] (22.center) to (37.center);
		\draw [thick] (10.center) to (12.center);
		\draw [thick] (20.center) to (23.center);
		\draw [thick] (24.center) to (9.center);
		\draw [thick, bend right=90, looseness=1.50] (27.center) to (30.center);
		\draw [thick] (5.center) to (16.center);
		\draw [thick, bend right=90, looseness=1.00] (32.center) to (34.center);
		\draw [thick] (10.center) to (7.center);
		\draw [thick] (11.center) to (28.center);
		\draw [thick] (9.center) to (26.center);
		\draw [thick] (20.center) to (6.center);
		\draw [thick] (24.center) to (26.center);
	\end{pgfonlayer}
\end{tikzpicture}